\documentclass[letterpaper]{article} % DO NOT CHANGE THIS
\usepackage{aaai2026}
\usepackage{times}  % DO NOT CHANGE THIS
\usepackage{helvet}  % DO NOT CHANGE THIS
\usepackage{courier}  % DO NOT CHANGE THIS
\usepackage[hyphens]{url}  % DO NOT CHANGE THIS
\usepackage{graphicx} % DO NOT CHANGE THIS
\usepackage{natbib}  % DO NOT CHANGE THIS AND DO NOT ADD ANY OPTIONS TO IT
\usepackage{caption} % DO NOT CHANGE THIS AND DO NOT ADD ANY OPTIONS TO IT
\usepackage{algorithm}
\usepackage{algorithmic}
\newcommand{\sfli}[1]{{\color{black}#1}}
\usepackage{newfloat}
\usepackage{listings}
\DeclareCaptionStyle{ruled}{labelfont=normalfont,labelsep=colon,strut=off} % DO NOT CHANGE THIS
\floatstyle{ruled}
\newfloat{listing}{tb}{lst}{}
\floatname{listing}{Listing}
\usepackage{bm}
\usepackage{amsmath}
\usepackage{amsfonts}  % 或 amssymb（二者通常已互相依赖）
\usepackage{booktabs}
\usepackage{multirow}
\usepackage{xcolor}
\usepackage{makecell} % 关键宏包
\usepackage{tcolorbox}

\title{Instant Preference Alignment for Text-to-Image Diffusion Models}
\author{
    Yang Li\textsuperscript{\rm 1}
    , Songlin Yang\textsuperscript{\rm 2}, Xiaoxuan Han\textsuperscript{\rm 1}, Wei Wang\textsuperscript{\rm 1}\thanks{Wei Wang is the corresponding author}, Jing Dong\textsuperscript{\rm 1}, Yueming Lyu\textsuperscript{\rm 3}, Ziyu Xue\textsuperscript{\rm 4}
}
\affiliations{
    \textsuperscript{\rm 1}New Laboratory of Pattern Recognition, CASIA, \textsuperscript{\rm 2}The Hong Kong University of Science and Technology\\
    \textsuperscript{\rm 3}Nanjing university, \textsuperscript{\rm 4}Academy of Broadcasting Science, NRTA

}

\usepackage{bibentry}
\begin{document}

\maketitle

\begin{abstract}
% change, 改，狠狠的改，要突出重点，围绕instant这件事情，只说他，不说别的！！！
% 和intro的形式其实应该保持相似，首先引入instant-preference alignment任务，接着简短说明当前方法对此任务的不适用，接着是我们提出的解决方案（解决方案是紧紧围绕在当前任务的）

Text-to-image (T2I) generation has greatly enhanced creative expression, yet achieving preference-aligned generation in a real-time and training-free manner remains challenging. Previous methods often rely on static, pre-collected preferences or fine-tuning, limiting adaptability to evolving and nuanced user intents. In this paper, we highlight the need for instant preference-aligned T2I generation and propose a training-free framework grounded in multimodal large language model (MLLM) priors. Our framework decouples the task into two components: preference understanding and preference-guided generation. For preference understanding, we leverage MLLMs to automatically extract global preference signals from a reference image and enrich a given prompt using structured instruction design. Our approach supports broader and more fine-grained coverage of user preferences than existing methods. For preference-guided generation, we integrate global keyword-based control and local region-aware cross-attention modulation to steer the diffusion model without additional training, enabling precise alignment across both global attributes and local elements. The entire framework supports multi-round interactive refinement, facilitating real-time and context-aware image generation. Extensive experiments on the Viper dataset and our collected benchmark demonstrate that our method outperforms prior approaches in both quantitative metrics and human evaluations, and opens up new possibilities for dialog-based generation and MLLM-diffusion integration.

\end{abstract}

\begin{figure}[!th]
    \centering
    \includegraphics[width=0.9\linewidth]{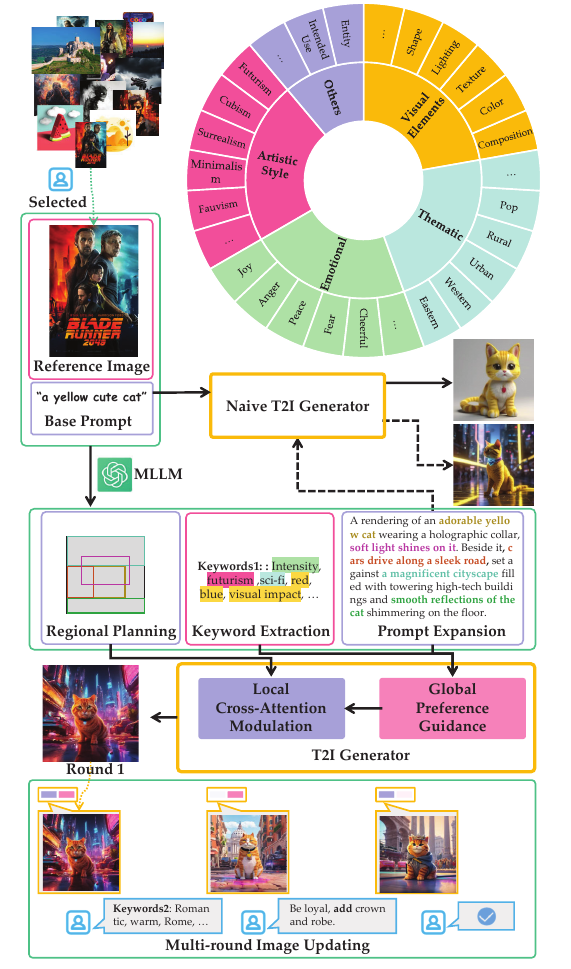}
    \vspace{-0.3cm}
    \caption{
    Overview of our instant preference-aligned T2I framework. It decouples the task into preference understanding and preference-guided generation, enabling training-free, real-time, multi-round generation.
    }
    \label{pic:teaser}
    \vspace{-0.5cm}
\end{figure}

 % We comprehensively understand preference using MLLMs priors. Then, several strategies are proposed to realize preference-guided T2I generation. Moreover, real-time refinements are supported for better alignment.

\section{Introduction}
\label{sec:intro}

% \[
% \begin{minipage}{0.9\linewidth}
% \centering
% \itshape
% ``One picture worth ten thousand words!''\\[1ex]
% -- Fred R.Barnard, 1921
% \end{minipage}
% \]

% 写作的路线整理一下：首先定义好instant preference是什么，强调在这其中有哪些方面是比较重要的
% 紧接着说明现在的方法（强化学习、或基于特征注入的方式），他们关注点是什么存在什么样的不足
% 最后一段， 先概括讲方法，接着在强调每一个模块设计（首先说明针对instant preference的问题，观察到的不足，然后我们提出什么样的方法，来解决这个问题，重点凸显出是围绕着instant preference的）

Text-to-image (T2I) generation has significantly boosted user creativity, yet efficiently achieving preference-aligned generation remains a pressing challenge.
Previous approaches rely on pre-collected data, assuming user preferences remain unchanged.
 % Previous approaches rely on static or pre-collected preferences, assuming user intent is stable and domain-specific.
However, user preferences inherently exhibit strong temporal dynamics and are multifaceted. Therefore, it is essential for T2I models to adapt dynamically to user preferences at inference time. In this paper, we highlight the need for instant preference-aligned generation.

% and context-specific locality
% 及时调整,及时调整
% instant preference 的挑战是什么，instant，强调出在实现instant 理解和instant生成两方面存在的问题
Typically, preference-aligned image generation involves two key components: \textit{\textbf{Preference Understanding,}} capture user preference signals with provided information, and \textit{\textbf{Preference-Guided Generation,}} guide image generation with the preference signals.  
% to ensure alignment with intended visual intent. 
However, instant preference-aligned generation imposes stricter requirements: 

(1) In the preference understanding stage, the system must efficiently infer multifaceted and nuanced preferences, even from minimal input (e.g., a single image).
Approaches like Viper~\cite{salehi2024viper} rely on labor-intensive human expert acquisition, which introduces a significant \sfli{delay}
% temporal overhead 
that hinders its adaptability to rapidly evolving preferences. Alternatively, other techniques~\cite{ruiz2023dreambooth, li2024beyond, wang2024instantstyle, salehi2024viper, shen2024pmg} leverage pretrained models for specific object or style preference extraction, yet they often \sfli{lack a comprehensive understanding of user preferences} and overlook critical contextual content, leading to poor preference alignment.
% yet they often lack comprehensive user preference coverage and overlook critical contextual content, leading to poor preference alignment.

(2) In the preference-guided generation stage, a major challenge lies in seamlessly integrating inferred preferences with contextual information without further fine-tuning. This complexity stems from preferences inherently influencing both global attributes (e.g., artistic style, emotional tone) and local elements (e.g., specific objects or themes) within an image. While Reinforcement Learning from Human Feedback (RLHF)~\cite{wallace2024diffusion, fan2024reinforcement, black2023training, zhang2024learning, xu2023imagereward, von2023fabric} has proven effective in aligning models with general human preferences, it requires additional training and relies on coarse-grained feedback signals, making it difficult to capture the nuanced influence of preferences on both global attributes and local details, even after extensive learning.

To tackle these challenges, we propose a training-free, instant-preference-aligned image generation framework with MLLM priors, encompassing both preference understanding and preference-guided generation. 

% Specifically, 
For the \textbf{\textit{understanding}} stage, unlike Viper~\cite{salehi2024viper}, which necessitates manual preference definition by expert users, we leverage MLLM to analyze reference images, extracting global keywords that reflect the image's overall tone, texture, and other salient features. 
% To enhance the discriminability of these keywords,
\sfli{To enhance the comprehensiveness of the keywords},
we introduce four critical categories of preference priors: artistic style, emotional/atmospheric resonance, thematic, and visual elements. As shown in Table~\ref{tab:keywords_alignment}, 
\sfli{these four categories cover the vast majority of user preferences.}
% these encompass a broad spectrum of preferences. 
Subsequently, the MLLMs are instructed to perform fine-grained enrichment for a given base prompt, \sfli{utilizing comprehensive keywords and context content from reference images to generate a detailed preference-aligned prompt.}
% , instantly yielding a context-aware and preference-aligned prompt. 

For~\textbf{\textit{generation}} stage, to facilitate seamless integration of preference features without additional training, we introduce the preference guidance at both global and local levels. Specifically, the global preference guidance is adopted to steer image generation using preference keywords, thereby maintaining coherence with the reference images. Furthermore, our local cross-attention modulation applies region constraints to modulate the attention mechanism, ensuring precise control over spatially grounded attributes, facilitating the placement and rendering of preference-relevant elements and visual details.

Crucially, the entire framework supports interactive feedback at each stage, enabling real-time, multi-round refinement, toward more precise and preference-aligned image generation. Extensive experiments demonstrate our method significantly outperforms comparison methods in both metrics and user studies on the Viper Dataset~\cite{salehi2024viper} and our collected dataset, achieving superior instant preference-aligned image generation capability. 

\textbf{Our contributions can be summarized as:}
\begin{itemize}
    \item We pioneer the instant preference alignment framework for real-time T2I generation, advancing dynamic adaptation to multifaceted and evolving preferences.
    \item We achieve preference understanding by integrating MLLMs with unique instruction design, thereby profoundly achieving broader and more fine-grained preferences than previous methods.
    \item For preference-guided generation, we eliminate the fine-tuning burden of prior approaches, achieving effective diffusion model control through a more effective global-local disentanglement at inference time.
    % \item We highlight the need for instant preference-guided T2I generation, and propose a training-free workflow to achieve real-time T2I preference alignment.
    % \item For preference understanding, we leverage MLLMs to automate the extraction process and design tailored instructions to achieve broader and more fine-grained user preferences than previous methods.
    % \item For preference-guided generation, we eliminate the fine-tuning burden required by previous methods and instead achieve effective control over the diffusion model through a more effective global-local disentanglement at the inference time.
    
    \item Extensive experiments demonstrate that our method outperforms previous approaches and supports multi-round interactive refinement, opening up new possibilities for dialog-based generation and the integration of LLMs with diffusion models.
\end{itemize}

\begin{figure*}[!h]
    \centering
    \includegraphics[scale=0.6]{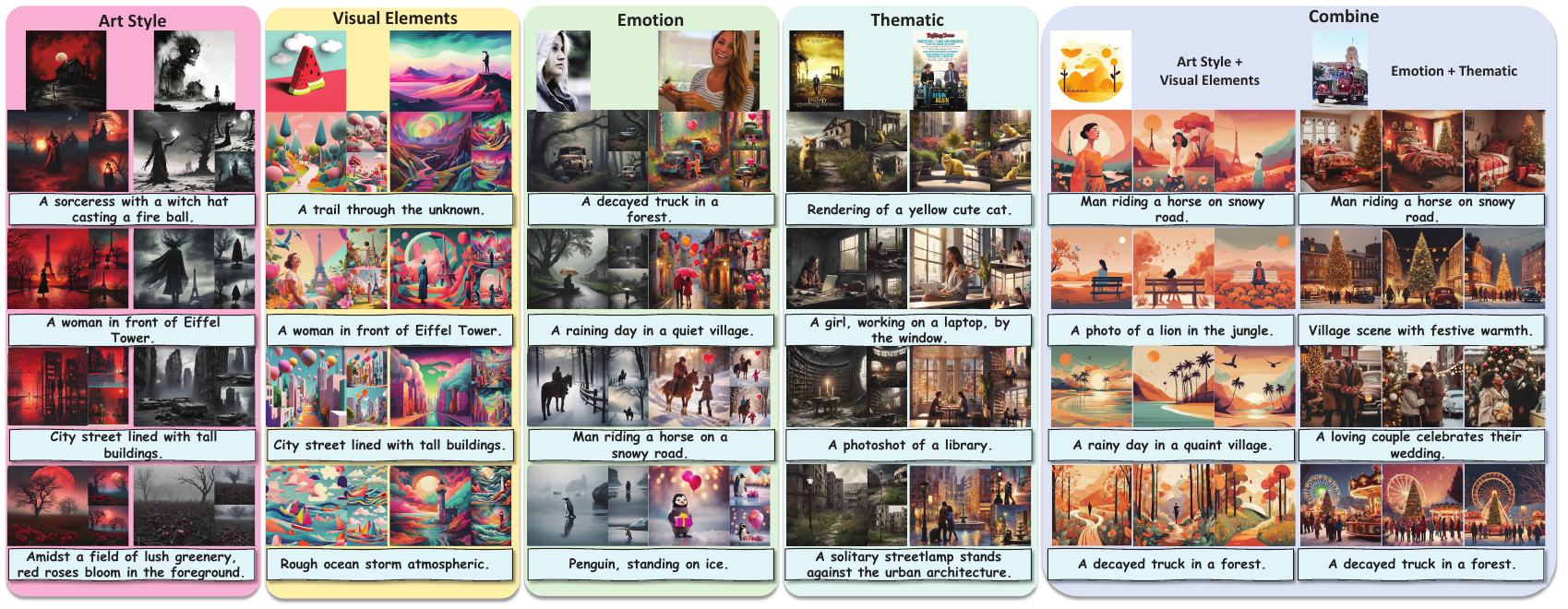}
    \vspace{-0.2cm}
    \caption{Examples of T2I generations aligned with preferences spanning art styles, visual elements, emotion, and thematic combinations are presented. Each image integrates preference features from reference images with base prompts. The successful incorporation of global attributes, such as color palettes, and additional entities further enhances alignment with the reference images. For instance, the image in the fourth row and second column distinctly showcases a black and gray color scheme alongside broken cars, exemplifying enhanced preference alignment.}
    \vspace{-0.4cm}
    \label{fig:main_result}
\end{figure*}

\section{Related Work}
\label{sec:related_work}

\subsection{Preference Alignment of T2I Generation} 
Generative models~\cite{rombach2022high} are typically trained on large-scale datasets, resulting in generally applicable image outputs. There has been growing interest in tailoring T2I models to better align with human preferences. Methods for aligning T2I outputs with user requirements can be broadly divided into two primary categories: 

\noindent{\textbf{Preference Learning.}} Methods~\cite{clark2023directly, prabhudesai2023aligning, fan2024reinforcement, black2023training} utilize reinforcement learning~\cite{fan2024reinforcement, black2023training} or direct preference optimization~\cite{wallace2024diffusion} to align T2I models with stable human preference by fine-tuning diffusion models using preference data pairs~\cite{xu2023imagereward, kirstain2023pickapic, wu2023human}. While these methods substantially improve the quality and aesthetics of generated images, they require extensive training data and necessitate large-scale retraining to adapt to new preferences. Other approaches incorporate binary feedback~\cite{von2023fabric} or ranking strategies~\cite{tang2023zeroth} during training, but they often yield suboptimal results due to their limited feedback information.

\noindent{\textbf{Customized Content Generation.}} The customized content generation aims to generate images aligned with user-specified content. Subject customization methods encode user-specified subjects into generated images via fine-tuning models~\cite{gal2022image, ruiz2023dreambooth, li2024beyond} or training an encoder~\cite{ye2023ip}. Meanwhile, style transfer~\cite{wang2024instantstyle, hertz2024style} focuses on learning image style from a reference image and then incorporating it into new artwork creation. Prompt Rewrite~\cite{chen2024tailored} leverages the Large Language Model to rewrite text prompts with user historical artistic style preferences. Similarly, Viper~\cite{salehi2024viper} explores visual preference alignment (i.e.,, color palette, vibe, and lighting) by collecting style attributes from expert comments, which lacks scalability for diverse preferences at scale. Collectively, existing methods primarily focus on either subjects or styles, neglecting the multifaceted nature of preferences (e.g., emotion, theme-related objects, style, and their mixture) during image generation. To bridge this gap, we propose leveraging MLLMs to automatically understand diverse preference signals from an informative user-preferred image and integrating real-time user feedback, enabling a more comprehensive and adaptable preference-aligned image generation.

\subsection{Spatial-Conditioned Image Generation}
These methods aim to improve the compositional generation ability of T2I models by using priors from segmentation maps, layouts, sketches, and etc. Getting it Right~\cite{chatterjee2025getting} seeks to retrain the diffusion model with a large-scale spatially focused dataset. Other methods~\cite{zhang2023adding, li2023gligen, avrahami2023spatext, wu2024ifadapter, mou2024t2i} mainly train additional modules into the diffusion model for spatial-condition injection. In contrast, training-free methods manipulate the inference phase by optimizing the cross-attention map~\cite{chen2024training, chefer2023attend} or image latents~\cite{lian2023llm, liu2022compositional} according to spatial or semantic constraints from labeled data or MLLM output~\cite{lian2023llm, yang2024mastering}. Despite their benefits, these training-free methods currently offer only coarse compositional control, leading to unsatisfactory image generation, especially when dealing with complex prompts.
\begin{figure*}[!th]
    \centering
    \includegraphics[width=0.98\linewidth]{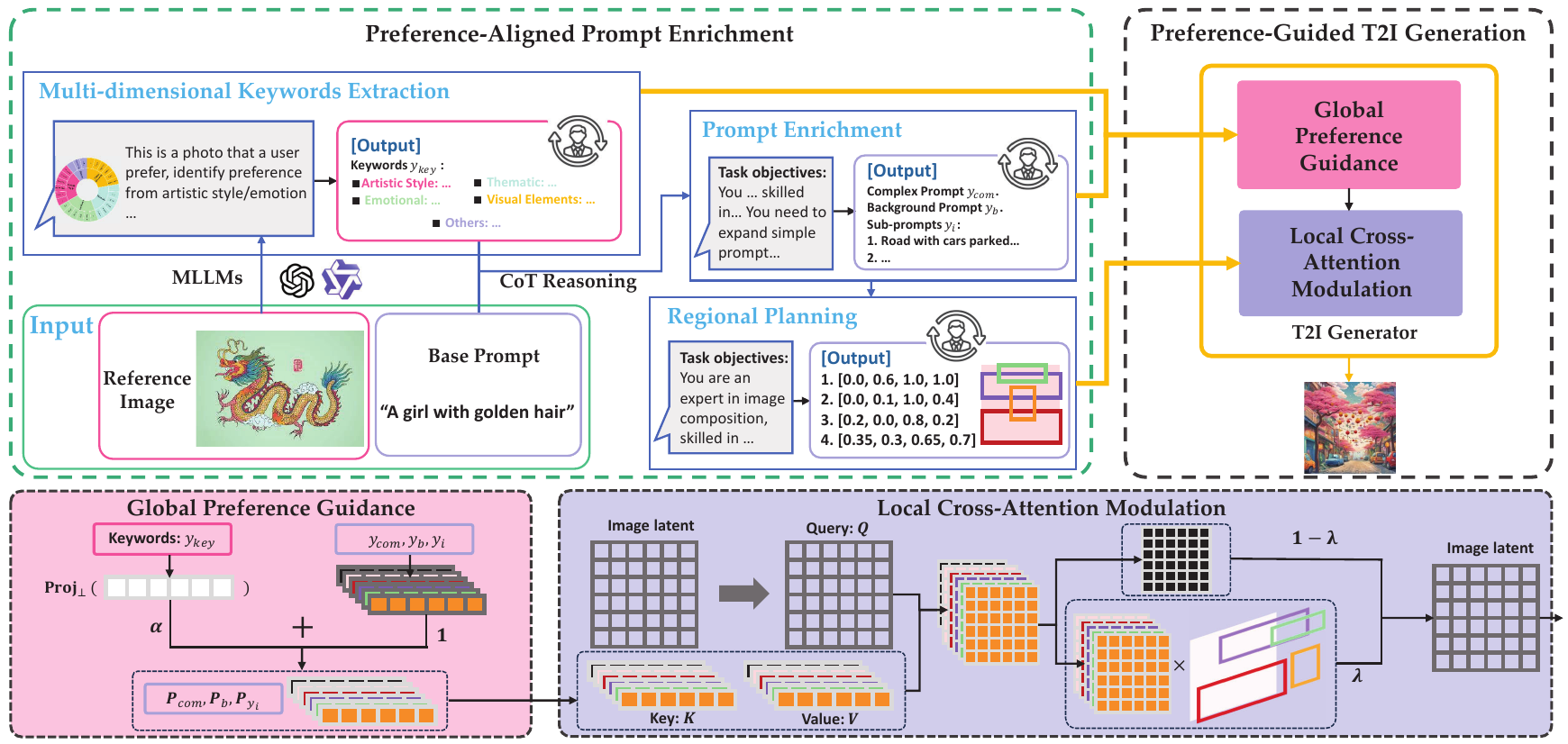}
    \vspace{-0.2cm}
    \caption{Overview of our preference-guided T2I framework. \textbf{(1) Preference-Aligned Prompt Enrichment}, which leverages MLLMs to extract keywords from reference images and enrich the entity description based on these keywords. \textbf{(2) Preference-Guided T2I Generation}, which uses global preference guidance for overall alignment, and applies local cross-attention modulation with regional constraints to facilitate fine-grained preference-aligned visual rendering.}
    \label{fig:pipeline}
    \vspace{-0.4cm}
\end{figure*}

\section{Method}
\label{sec:method}

Our preference-guided T2I framework can be summarized as: \textit{\textbf{(1) Understanding: Preference-Aligned Prompt Enrichment.}} Given preferred reference images (or other information), the goal is to enrich a prompt that aligns with the multifaceted preferences. This process involves extracting key preference features from the reference image and enriching the base prompt to reflect the multiple preference contexts while maintaining the semantic meaning of the base prompt. \textit{\textbf{(2) Generation: Preference-guided T2I.}} Using the preference-aligned prompt, the T2I model generates an image that not only reflects the global attributes but also exhibits the preference-aligned local elements described in the enriched prompt.

\subsection{Preliminary}
\label{diffusion}
\noindent\textbf{Diffusion-Based T2I Generation.}
The Stable Diffusion Model~\cite{rombach2022high, podell2023sdxl}, which integrates CLIP Text Model~\cite{radford2021learning} $\mathcal{T}_{text}(\cdot)$, VAE~\cite{razavi2019generating}, and a latent U-Net~\cite{ho2020denoising} $\epsilon_{\theta}(\cdot)$, plays a crucial role in text-to-image generation. Given any user-provided prompt $y$, the latent U-Net denoiser $\epsilon_{\theta}(\cdot)$ is trained as:
\begin{equation}
    \mathcal{L}=\mathbb{E}_{\bm{x}\sim \mathcal{E}(\bm{I}), \bm{\epsilon}\sim\mathcal{N}(0, 1), t}\left [ \left \| \epsilon - \epsilon_{\theta}(\bm{x}_t, t, \mathcal{T}_{text}(y))\right \| \right ],
\end{equation}
where $t$ is the timestep, $\epsilon$ denotes for the unscaled noise, and $\bm{x}_{t}$ is the noised latent. 

\noindent\textbf{Cross-Attention for Text Condition.} The latent image feature $\bm{z}$ is updated based on the prompt $y$ through cross attention module as follows:
\begin{equation}
    \bm{Q} = \bm{W}^q\bm{z}, \bm{K} = \bm{W}^k\mathcal{T}_{text}(\bm{y}), \bm{V} = \bm{W}^v\mathcal{T}_{text}(\bm{y}),
\end{equation}
\vspace{-0.5cm}
\begin{equation}
    \text{Attention}(\bm{Q},\bm{K}, \bm{V})=\text{Softmax}(\frac{\bm{Q}\bm{K}^T}{\sqrt{d}})\bm{V}.
\end{equation}
where $d$ is the output dimension of \textbf{K}ey and \textbf{Q}uery, $\bm{W}^q$, $\bm{W}^k$, and $\bm{W}^v$ map the inputs to \textbf{Q}uery, \textbf{K}ey, and \textbf{V}alue features, respectively.

\subsection{Preference-Aligned Prompt Enrichment}
\label{sec: prompt_guidance}

This process can be divided into two steps: Firstly, using the MLLMs to extract the keywords from the preferred reference images. Secondly, enriching the given prompt based on these preference keywords. 

\noindent\textbf{Multi-dimensional Keywords Extraction.} 
% To caption the keywords more comprehensively and accurately, 
We systematically classify preference content into Artistic Style, Emotional/Atmospheric Resonance, Thematic, Visual Elements, and other hard-to-classify preferences. This enables the extraction of a richer, tailored signal that integrates both high-level concepts and detailed attributes for an enhanced user experience. Let $X_{p}$ be the querying image that a tested user would prefer. We then leverage a pre-trained MLLM's strong image understanding and multimodal chain-of-thought reasoning ability~\cite{zhang2023multimodal} to imitate expert perception of querying images and extract preference keywords:
\begin{equation}
    y_{\text{key}}=\text{MLLM}_{\text{ext}}(X_{p}).
\end{equation}

\noindent\textbf{Keyword-Conditioned Prompt Enrichment.} For a target base prompt, different preferences would require distinct descriptions. Therefore, we instruct the MLLM to add contextually-interactable entities that embody the desired preferences. This process generates a complex prompt, denoted as $y_{\text{com}}$, comprising a set of $n$ entities, $\left \{ E_{i}\right \}^{n}_{i=0}=\left \{E_{0}, E_{1}, ..., E_{n}\right \}$. Inspired by RPG~\cite{yang2024mastering}, the MLLM simultaneously refines attribute descriptions for each entity, driven by preference keywords, to enhance their preference alignment. This process is formulated as:
\begin{equation}
    \left \{y_{i}\right \}^{n}_{i=0}=\left \{ y_{0}, y_{1}, ..., y_{n}\right \} = \text{MLLM}_{\text{exp}}(\left \{ E_{i}\right \}^{n}_{i=0}, y_{\text{key}}),
\end{equation}
$\left \{y_{i}\right \}^{n}_{i=0}$ are the sub-prompts for each objects. We also generate a background prompt $y_{\text{b}}$ that does not contain any objects to define the background information. Finally, these are merged with the complex prompt $y_{\text{com}}$ to form the expanded prompt group: $\left \{y_{\text{com}}, \left \{y_{i}\right \}^{n}_{i=0}, y_{\text{b}}\right \}$.

\subsection{Preference-Guided T2I Generation}
\label{subsec: image_generation}
% In the previous section, we discussed our perspective of splitting the preference visual content generation as preference entity and preference keywords. Here, we will use these two preference-related sub-parts to generate the user-preferred image with a specified \textit{subject}.
% In this section, we use expanded prompts for T2I generation. For the preference entity combination and generation, we use \textbf{Regional Planning} with the help of a MLLM to enrich the entity and supposed location in the generated image. Then, the \textbf{Cross-Attention Re-Organization} and \textbf{Global Preference Guidance} are deployed to manage the final generation. 

In this section, we apply the enriched prompt group for T2I generation. To better guide the model with preference signals, we first employ \textit{\textbf{Global Preference Guidance}} to ensure alignment with global attributes. Subsequently, \textit{\textbf{Regional Planning}} and \textit{\textbf{Local Cross-Attention Modulation}} are utilized to constrain the layout and generation of specific entities, ensuring adherence to the desired visual details.

% 增加这部分描述，突出为了我们相关任务我们做出的改进，后面的local cross也是如此
\noindent\textbf{Global Preference Guidance.} We encode the extracted preferences keywords, $y_{\text{key}}$, from each preferred image into the text space using the text encoder $\mathcal{T}_{text}(\cdot)$. As illustrated in Fig.~\ref{fig:pipeline}, this encoded preference embedding is added to each prompt in $\left \{{y_{\text{com}}, \left \{y_{i}\right \}_{i=0}^{n}, y_{\text{b}}}\right \}$ to steer generation towards the preferences of each instance:
\begin{equation}
    \bm{P}_{y_{i}} = \mathcal{T}_{text}(y_{i}) + \alpha \cdot\text{Proj}_{\perp}(\mathcal{T}_{text}(y_{\text{key}})),
\end{equation}
where $\text{Proj}_{\perp}$ projects the preference embedding orthogonally to each prompt embedding, with $\alpha$ controlling the strength of the preference modification. Unlike Viper~\cite{salehi2024viper}, such a projection operation significantly enhances preference alignment while robustly preserving the original prompt's context. This process gives us the modified embedding group $\left \{{\bm{P}_{\text{com}}, \left \{\bm{P}_{y_{i}}\right \}_{i=0}^{n}, \bm{P}_{\text{b}}}\right \}$. The comparison between our method and Viper is outlined in \textcolor{black}{Supp.~Sec.~C}.

\noindent\textbf{Regional Planning.} All sub-prompts $\left \{y_{i}\right \}_{i=0}^{n}$ are then feed into the MLLM to assign layouts. Since directly querying the MLLM for layout information can lead to inaccuracies, we provide manually curated in-context examples to enhance bounding box prediction accuracy. Specifically, for an image of size $H \times W$, such process assigns corresponding bounding boxes $\left \{B_i\right \}^{n}_{i=0} = \left \{ B_{0}, B_{1}, ..., B_{n}\right\}$ for each sub-prompt.
Each bounding coordinate is normalized to $[0,1]$, within the format $\left \{x_\text{left}, y_\text{top}, x_\text{right}, y_\text{bottom}\right \}$. Regions not covered by sub-region boxes are considered background. To mitigate occlusion from overlapping boxes, particularly for smaller entities, we sort the generated sub-region boxes and their corresponding embedding in descending order of area.

% 这里的优化主要表述，修改local entity这部分如何和prefrence之间有绑定，同时在后面的实验部分，我们也应该提供相应的支持
% 总结为：第一：preference align的sub-prompt更完美解释；第二：则是为了确保与preference align的多个物体都可以出现，在这里我们就是通过分别强调nuanced visual details and interrelationships
% 在与RPG的对比上抓住重点，resize策略是一个很差的策略会导致严重的blurring以及和语义不相符，尤其是对那些存在明显overlapping的区域

\noindent\textbf{Local Cross-Attention Modulation.} Even with recent advancements like RPG, diffusion models often struggle to accurately interpret complex prompts, particularly when the nuanced visual details of specified entities and their interrelationships must precisely align with user preferences. To address this challenge, we blend multi-entities, $\left \{ E_{i}\right \}^{n}_{i=0}$, into a unified image using a modified cross-attention mechanism, as shown in Fig.~\ref{fig:pipeline}. Specifically, at each timestep $t$, within each cross-attention layer, the embeddings from $\left \{{\bm{P}_{\text{com}}, \left \{\bm{P}_{y_{i}}\right \}_{i=0}^{n}, \bm{P}_{b}}\right \}$ are passed in parallel to generate generate corresponding latent image features $\bm{z}_{t}^{\text{com}}$, $\left \{\bm{z}^{i}_{t}\right \}_{i=0}^{n}$, and $\bm{z}_{t}^{\text{b}}$. This process can be formalized as follows:
\begin{equation}
    \bm{z}_{t}^{i}=\text{Attention} \left (\bm{W}^{q}\bm{z}_t, \bm{W}^{k}\bm{P}_{i}, \bm{W}^{v}\bm{P}_{i}\right ),
    % \bm{z}_{t-1}^i=Softmax(\frac{\bm{W^q} \phi(z_t)(\bm{W^k} \mathcal{T}_{text}(\hat{y_{i}}))}{\sqrt{d}})(\bm{W^v} \mathcal{T}_{text}(\hat{y_{i}})),
\end{equation}
where $\bm{z}_t$ is the input latent image feature.
Then, we replace the generated latent image feature of $\bm{z}_{t}^{\text{b}}$ with $\left \{\bm{z}^{i}_{t}\right \}_{i=0}^{n}$ according to the sub-region boxes $\left \{B_i\right \}^{n}_{i=0}$, enabling spatial control over each object. The process can be described as:
\begin{align}
    \bm{z}^{\text{b}}_{t} &= \bm{M}_{B_{i}} \cdot \bm{z}^{i}_{t} + (1-\bm{M}_{B_{i}}) \cdot \bm{z}^{\text{b}}_{t}, \\
    \bm{z}^{\text{C}}_{t} &= \bm{z}^{\text{b}}_{t},
\end{align}
% \begin{align}
%     \bm{z}^{\text{b}}_{t-1}(B_{i}) &= \bm{z}^{i}_{t-1}(B_{i}), \\
%     \bm{z}^{\text{C}}_{t-1} &= \bm{z}^{b}_{t-1},
% \end{align}
where $\bm{M}_{B_{i}}$ is a mask indicating regions within bounding box $B_{i}$. The replacement operation follows the descending order specified by \textbf{Regional Planning}.
To ensure seamless and harmonious blending of the background and objects, we employ a weighted sum of the composed latent image feature $z^{\text{C}}_{t}$ and the complex latent image feature $z^{\text{com}}_{t}$ to output the final latent image feature $z_{t}$.
\begin{equation}
    \bm{z}_{t}=\lambda \cdot \bm{z}^{\text{com}}_{t} + (1-\lambda)\cdot \bm{z}^{\text{C}}_{t}.
\end{equation}
$\bm{z}_{t}$ is the updated latent image feature passed to the next module, and $\lambda$ is a hyper-parameter balancing individual entity contribution and complex prompt.

% \subsection{Application: Multi-Round Image Updating}
% % 需要修改，变成与method相匹配的表示
% Both quantitative and qualitative evaluations demonstrate the superior performance of our method, even without iterative user feedback. Crucially, our workflow inherently further supports multi-round image refinement, a vital capability for real-world instant preference alignment. This includes real-time adjustments to extracted keywords, prompt content, and layout planning. Fig.~\ref{fig:interactive_caption} showcases detailed examples of real-user interactions with our system, illustrating how generated images achieve full alignment with user preferences. Refer to \textcolor{red}{Supp} for more results.

\begin{figure*}[t]
    \centering
    \includegraphics[width=0.95\linewidth]{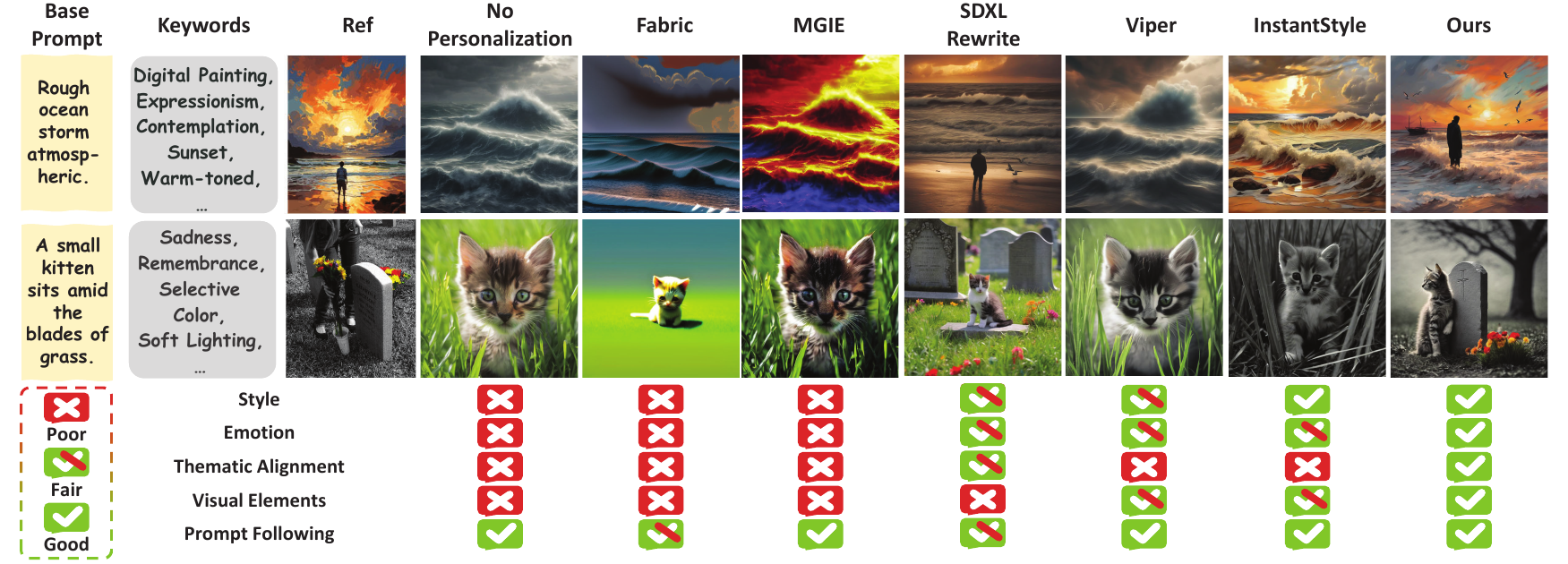}
    \vspace{-0.2cm}
    \caption{Comparison with baselines. Neither Fabric nor MGIE is capable of generating preference-aligned results. While SDXL Rewrite can produce images close to the reference, it struggles with consistent generation from complex prompts and fails to preserve preference signals. Viper fails to achieve thematic alignment and may lose style and emotion-related elements. InstantStyle solely focuses on transferring reference colors and styles. Finally, our method generates images that adhere to the base prompt while embedding preference signals closely aligned with the reference images.}
    \label{fig:compare}
    \vspace{-0.2cm}
\end{figure*}

\section{Experiments}
\label{sec:experiments}

\begin{table*}[ht]
    \caption{Quantitative comparison with other methods.
    % We evaluate from style, emotional, base prompt alignment, and keywords alignment. 
    % Our method demonstrates overall optimal performance, consistently outperforming other approaches. We also include human and GPT-4o~\cite{hurst2024gpt} to evaluate the matching between preferred reference images and generated images. 
    \sfli{Our approach achieves optimal overall performance while demonstrating significant advantages in preference studies.}
    The number indicates the percentage of participants who vote for the result. The best results are in highlighted \textbf{bold} and the second-best ones are \underline {underlined}.}
    \footnotesize
    \centering
    \vspace{-0.2cm}
    \begin{tabular}{cccccccc}
    
    \toprule[1pt]
        \multirow{2}{*}{Methods} & \multicolumn{4}{c}{Metrics$\uparrow$} & & \multicolumn{2}{c}{User Preference Study$\uparrow$} \\
         \cline{2-5}\cline{7-8}
         & 
        Style Loss & Emo Acc & CLIP Score & ImageReward & & Human as Evaluator & GPT-4o as Evaluator \\
        \midrule
        NP & 0.569 & 31.1$\%$ & \textbf{0.295
} & 4.93$\%$ & & 2.21$\%$ & 2.24$\%$ \\
        % Zero-RankSGD~\cite{tang2023zeroth} & 0.569 & 31.2$\%$ & 0.274 & 5.02$\%$ & & 2.32$\%$ & 2.79$\%$ \\
        Fabric & 0.556 & 28.0$\%$ & 0.245 & 1.31$\%$ & & 2.04$\%$ &  2.17$\%$ \\
        MGIE & 0.593 & 33.4$\%$ & 0.283 & 5.56$\%$ & & 2.93$\%$ & 3.09$\%$ \\
        SDXL Rewrite & 0.635 & \underline{44.2$\%$} & 0.273 & 22.3$\%$ & & 9.92$\%$ & 11.3$\%$ \\ 
        % SD3-L Rewrite & 0.635 & 42.7$\%$ & 0.274 & \underline{\textbf{24.1$\%$}} &  & \underline{\textbf{18.6$\%$}} & \underline{\textbf{25.9$\%$}} \\ 
        Viper & 0.647 & 42.5$\%$ & 0.282 & \underline{23.6$\%$} &  & \underline{15.8$\%$} & \underline{22.4$\%$} \\
        InstantStyle & \textbf{0.712} & 36.6$\%$ & 0.252 & 16.9$\%$ & & 13.9$\%$ & 14.5$\%$ \\ 
        Ours & \underline{0.663} & \textbf{60.2$\%$} & \underline{0.288} & \textbf{25.4$\%$} & & \textbf{53.2$\%$} & 
\textbf{44.3$\%$} \\
        \bottomrule[1pt]
    \end{tabular}

    \label{tab:multi_metric}
    \vspace{-0.4cm}
\end{table*}

\begin{table}[!ht]
    \centering
    \tiny
    \caption{Comparing with Viper on the Viper dataset.}
     \vspace{-0.2cm}
    \begin{tabular}{ccccccc}
    \toprule
               & \makecell{Style\\Loss} & \makecell{Emo\\Acc} & \makecell{CLIP\\Score} & ImageReward & Thematic & \makecell{Visual\\Elements} \\
    \midrule[0.1pt]
        Viper & 0.69 & 38.7$\%$ & 0.276 & 19.8$\%$ & 6.01 & 5.84 \\
        Ours & \textbf{0.77} & \textbf{56.1$\%$} & \textbf{0.278} & \textbf{80.2$\%$} & \textbf{7.12} & \textbf{5.97} \\
    \bottomrule
    \end{tabular}
    \label{tab:viperdataset}
    \vspace{-0.3cm}
\end{table}

\subsection{Experimental Setting}
\noindent\textbf{Implementation Details.} \textbf{\textit{(1) Target Models: }}Our pipeline is extensible, allowing the integration of diverse MLLMs and Diffusion backbones, as demonstrated in the \textcolor{black}{Supp Sec.~D}. In the main paper, we choose Qwen-VL-72B and Stable Diffusion XL as the backbones.
%%% can delete
We carefully design the question templates and in-context examples to trigger the CoT planning for the \textbf{Keywords Extraction}, \textbf{Prompt Enrichment}, and \textbf{Regional Planning}. The details of the templates and in-context examples are provided in the \textcolor{black}{Supp Sec.~A}.
For SDXL, we set the classifier guidance $\omega=5$ and use the DPM-Solver~\cite{lu2022dpm} with $30$ steps for sampling. \textbf{\textit{(2) Hyper-Parameters: }}We set the hyper-parameters $\lambda$ and $\alpha$ to $0.2$ and $0.7$, respectively. \textbf{\textit{(3) Test Data: }}We collect image data from four categories: culture, art, emotion, and movie data. The emotional pictures are selected from the large-scale visual emotion dataset EmoSet~\cite{yang2023emoset}, while the other images are downloaded from the Internet. An illustration of the dataset can be found in the \textcolor{black}{Supp Sec.~A}. Moreover, we also include the dataset from Viper~\cite{salehi2024viper} to further evaluate the generalization of our framework.

\noindent\textbf{Baseline Selection.} We aim to generate images that align with the preferences embedded in the user-provided images. To this end, we compare six State-of-the-Art methods:
    \textbf{\textit{(1) No Preference (NP).}} The original base prompt is used to generate images without any additional information.
    \textbf{\textit{(2) Fabric~\cite{von2023fabric}.}} This process iteratively updates the image generation. In each iteration, the new image is conditioned on user-selected liked and disliked images from the generated image pool. This process highlights that, in the absence of fine-grained preference signals, the generation process is prone to overfit, leading to suboptimal results.
    \textbf{\textit{(3) MGIE~\cite{fu2023guiding}}.} The method utilizes an MLLM to refine the user's original editing instructions, ensuring that the modified instructions guide the image editing process toward results aligned with the user's intent. Our comparison demonstrates that simple object attribute editing fails to align with the preference effectively.
    \textbf{\textit{(4) Prompt Rewrite~\cite{chen2024tailored}.}} We prompt the SDXL with the complex prompt $y_{\text{com}}$. This complex prompt ensures the integration of multiple preferred attributes with the base prompt and serves to evaluate the effectiveness of \textbf{Preference-Guided T2I Generation}.
    % \item \textit{DDPO~\cite{wallace2024diffusion}.} We use their official checkpoint from Huggingface\footnote{https://huggingface.co/mhdang/dpo-sdxl-text2image-v1/tree/main}.
    % \item \textit{Zero-RankSGD~\cite{tang2023zeroth} (ZO-Rank).} Users provide ranked feedback on a selection of images, which is then used to refine the image generation.
    \textbf{\textit{(5) Viper~\cite{salehi2024viper}.}} Viper utilizes a collection of expert comments to summarize visual preferences. It modifies prompt embeddings and Classifier-Free Guidance to incorporate preference signals. We integrate MLLM-extracted keywords with their preference injection module to evaluate the effectiveness of our \textbf{Preference-Guided T2I Generation} in comparison to theirs.
    \textbf{\textit{(6) InstantStyle~\cite{wang2024instantstyle}.}} It is adopted to show that merely controlling specific aspects of preference (i.e., the color tone or style) is insufficient to represent user preferences comprehensively.

\noindent\textbf{Evaluation Metrics.} We evaluate preference using a suite of metrics, including style consistency, emotion accuracy, base prompt alignment, and keyword alignment. Style consistency is assessed using Style Loss~\cite{frenkel2025implicit}. An Emotion Classifier~\cite{yang2024emogen} is employed to measure the emotion alignment between the generated and the reference images. CLIP Score~\cite{radford2021learning} and ImageReward~\cite{gal2022image} are adopted for assessing the text-image alignment, where the text corresponds to the base prompt and the extracted keywords, respectively.
Following recipes in prior work~\cite{lee2024prometheus, yu2024few} that use MLLM-as-a-judge, we use GPT-4o to evaluate \textbf{\textit{Thematic}} and \textbf{\textit{Visual Elements}} alignment.
Furthermore, we conducted user studies to validate the preference alignment. Over 20 participants evaluated approximately $\sim$200 generated images, answering roughly $\sim$80 questions each to assess alignment with reference images.

\begin{table}[t]
    \tiny
    \centering
    \caption{We \sfli{design a user study} to assess the preference coverage across five categories and the alignment of keywords generated by MLLMs with \sfli{user} preferences.}
    \vspace{-0.2cm}
    \begin{tabular}{cccccc}
    \toprule[1pt]
        & \makecell{Artistic\\Style} & \makecell{Emotional} & \makecell{Thematic} & \makecell{Visual\\Elements} & Others \\
        \midrule[0.1pt]
       \makecell{PreferenceCoverage\\(Ratio$\%$)} & 26.5$\%$ & 22.9$\%$ & 21.4$\%$ & 24.9$\%$ & 4.3$\%$ \\
         \makecell{Alignment with Users\\(Accuracy$\%$)} & 82.3$\%$ & 85.2$\%$ & 80.4$\%$ & 84.7$\%$ & $-$ \\
    \bottomrule[1pt]
    \end{tabular}
    \vspace{-0.6cm}
    \label{tab:keywords_alignment}
\end{table}

\subsection{Comparison with Baselines}

\textbf{Qualitative Comparison.} Fig.~\ref{fig:compare} presents visual comparisons. Fabric often yields collapsed images because its human feedback is information-sparse and binary. While SDXL Rewrite can roughly reproduce color palettes and themes of the reference image, it struggles to precisely interpret complex prompts and may omit critical visual elements (e.g., the ``ship'' in the first row). Viper consistently fails to align generated images with preference signals and cannot seamlessly handle local visual details. InstantStyle primarily focuses on image colors or textures, struggles to evoke similar emotions or maintain thematic consistency with reference images. In contrast, our pipeline not only captures the global preference but also integrates preference-related entities and visual details, generating images that evoke a multifaceted impression similar to reference images while adhering to the base prompt. Please refer to Fig.~\ref{fig:main_result} for additional examples of preference-aligned generation using ours.
% The user study in the following could further support our analysis here, please refer to Tab~\ref{tab:multi_metric}.

% \begin{figure}[!ht]
%     \centering
%     \includegraphics[width=0.95\linewidth]{pics/keywords_alignment_crop.pdf}
%     \caption{We use user studies to assess the preference coverage across five categories and the alignment of keywords generated by MLLMs with users' preferences.}
%     \label{fig:keywords_alignment}
%     \vspace{-0.4cm}
% \end{figure}

\noindent\textbf{Quantitative Comparison.} \textit{(1) Preference Understanding:} We conduct a user study to assess the alignment of MLLM-generated preference keywords with user preferences. Participants are presented with a set of images, asked to select their preferred one, and then specify their preference reasons from: artistic style, emotional resonance, thematic, visual elements, and other reasons. This distribution of these selections (Fig.~\ref{tab:keywords_alignment}, first row) confirms the completeness of our categories. Notably, with only $4.3\%$ of users choosing ``Others''. Furthermore, the MLLM-generated keywords achieve over $80.0\%$ alignment accuracy with user preferences, as determined by direct user feedback on keyword relevance to their selections.
% This metric is determined by directly asking users whether the generated keywords match their opinions on their selections.
\textit{(2) Image Generation:} We calculate the Style Loss, Emotion Accuracy, Clip Score, and ImageReward to compare our method with the others.
\sfli{As shown in Table~\ref{tab:multi_metric}, our approach achieves an excellent trade-off between preference alignment and semantic preservation, leading to optimal overall performance. NP, a vanilla text-to-image model, slightly outperforms in CLIP Score but significantly neglects preference alignment.  InstantStyle, designed for style-customized generation and requiring additional training, excels in Style Loss but exhibits a clear performance gap compared to our method in other aspects.}
% As shown in Tab.~\ref{tab:multi_metric}, our method consistently achieves overall higher scores than comparison methods, demonstrating superior alignment with preferred images.
% Notably, our approach incurs only a negligible reduction in CLIP Score compared to NP, indicating an excellent trade-off between preference alignment and semantic preservation. While InstantStyle performs well in style alignment, its reliance on additional training and singular focus on style often leads to poor performance across other preference dimensions.
% SDXL Rewrite achieves competitive results with Viper as they either focus on thematic alignment or perform well in style alignment.
\sfli{
SDXL Rewrite and Viper are the closest works to our task, and our method exhibits clear advantages.
Moreover, quantitative results on the Viper dataset (Table~\ref{tab:viperdataset}) demonstrate superior performance over Viper and underline the generalization ability of our method.}

\begin{table}[t]
    \small
    \centering
    \caption{
    % Users are provided with two different images and asked to select the one they prefer. 
    \sfli{Users are presented with pairs of images for preference selection.}Then, they are shown groups of images generated from the two different images and asked to pick the one they prefer. The table shows whether the preference signal evoked by generated images \sfli{still aligns with the preferred images.}}
    % aligns with the reference image.}
     \vspace{-0.2cm}
    \begin{tabular}{ccc}
    \toprule[1pt]
         &  Human as Evaluator & GPT-4o as Evaluator \\
    \midrule[0.05pt]
    SDXL Rewrite & 68.3$\%$ & 64.2$\%$ \\
      Ours  &  \textbf{89.1$\%$} & \textbf{85.7$\%$} \\
    \bottomrule[1pt]
    \end{tabular}
    \vspace{-0.3cm}
    \label{tab:in_eval}
\end{table}

\begin{figure}[t]
    \centering
    \includegraphics[width=0.95\linewidth]{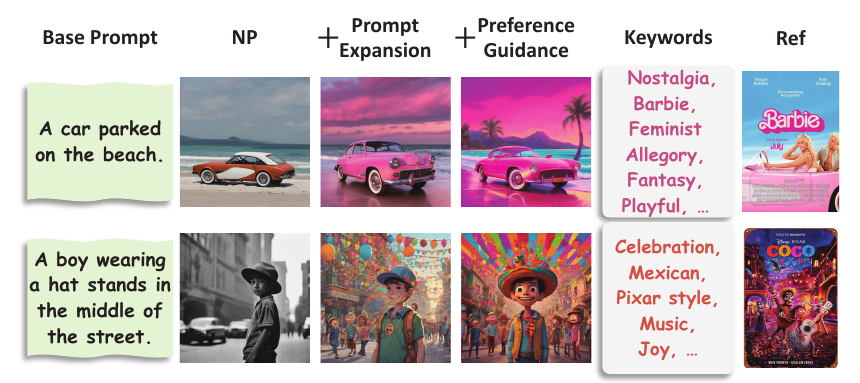}
    \vspace{-0.3cm}
    \caption{Global Preference Guidance further enhances the alignment of extracted keywords and generated images, e.g., the hat turned to ``Mexican Sombrero'' in the second row. The preference keywords are listed on the right.}
    \label{fig:abl_visual}
    \vspace{-0.6cm}
\end{figure}

\subsection{User Study}
We conduct user studies to evaluate the effectiveness of our pipeline with both human users and GPT-4o~\cite{hurst2024gpt}, performing two types of tests. (1) Users compared images generated by our method against those from other approaches, judging which better captured the essence of the reference image across various dimensions (art style, emotion, theme, visual elements, and others). Our method received nearly twice as much user preference as Viper, as shown in Table~\ref{tab:multi_metric}. (2) Users first selected their preferred image from a pair of distinct images. They were then presented with corresponding generated image groups and asked to choose their preference, assessing whether the preference signals evoked by the generated images aligned with their chosen reference. Our method achieved higher scores in this evaluation, as detailed in Table~\ref{tab:in_eval}.

\subsection{Ablation Study} 
% 这里对global的作用解释不够，需要再丰富一点
\noindent\textbf{Global Preference Guidance.}
We ablate the effect by incrementally incorporating \textbf{Prompt Enrichment} and \textbf{Global Preference Guidance}. Fig.~\ref{fig:abl_visual} depicts the results that Global Preference Guidance can further enhance the alignment of extracted keywords and generated images. Further comparison of it with Viper and the ablation of parameter $\alpha$, please refer to \textcolor{black}{Supp Sec.~C}.

\begin{figure}[t]
    \centering
    \includegraphics[width=0.95\linewidth]{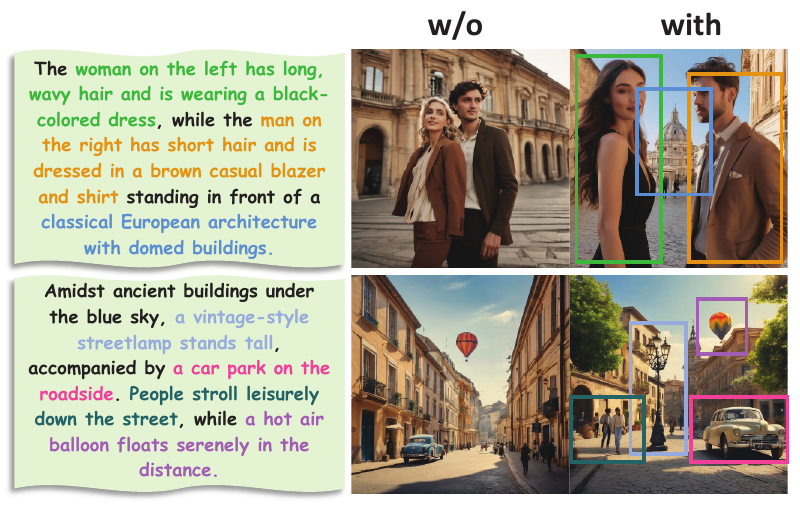}
    \vspace{-0.3cm}
    \caption{Ablation study of Local Cross-Attention Modulation. SDXL struggles to generate images aligned with long text, exhibiting attribute obfuscation and entity omission.
    % Our method effectively mitigates them.
    }
    \label{fig:abl_com}
    \vspace{-0.3cm}
\end{figure}

\noindent\textbf{Local Cross-Attention Modulation.}
When presented with complex prompts, SDXL often struggles to accurately render preference-related visual attributes and their corresponding entities, frequently yielding inadequate content, as shown in Fig.~\ref{fig:compare}. In Fig.~\ref{fig:abl_com}, our strategy, conversely, facilitates the robust composition of multiple entities by utilizing modified cross-attention, directly addressing this limitation.
% This strategy is employed to enable entity presence, and its effectiveness is validated through experiments, with results depicted in Fig.~\ref{fig:abl_com}. When complex prompts are used, SDXL struggles with the long prompts, often generating inadequate content, a phenomenon also observed in  Fig.~\ref{fig:compare}. However, our strategy facilitates the composition of multiple entities by utilizing modified cross-attention.

\subsection{Application: Multi-Round Image Updating}
% \label{sec:application}
% \begin{figure}
%     \centering
%     \includegraphics[width=0.95\linewidth]{pics/interactive_crop.pdf}
%     \caption{The Results of Multi-Round Image Evolving. Users can first generate their personalized results using our pipeline, Step \textbf{\normalsize{\textcircled{\scriptsize{1}}}\normalsize}. Then, the \textbf{Visual Preference Redirection} (step \textbf{\normalsize{\textcircled{\scriptsize{2}}}\normalsize}) and the \textbf{Entity Substitution} (step \textbf{\normalsize{\textcircled{\scriptsize{3}}}\normalsize}) can be used to refine the generated results.}
%     \label{fig:multi_round}
% \end{figure}

\begin{figure}[t]
    \centering
    \includegraphics[width=0.95\linewidth]{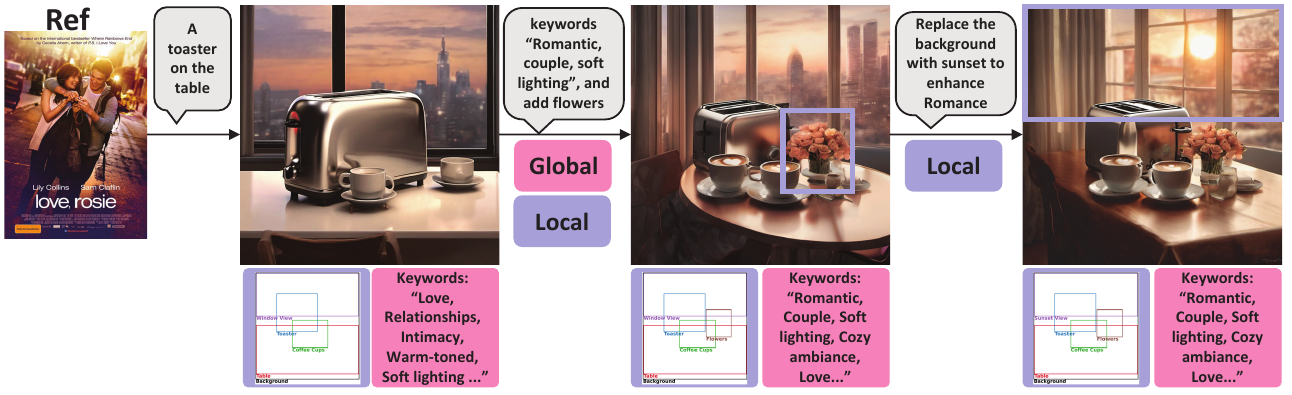}
    \vspace{-0.2cm}
    \caption{Interaction with real user and MLLMs.}
    \label{fig:interactive_caption}
    \vspace{-0.6cm}
\end{figure}

% Our pipeline also supports the updating of generated content to meet evolving individual preferences. We denote the multi-round image evolving from two aspects: (1) \textbf{Global Preference Changing}, interactively updates global preference according to the individual's requirement. (2) \textbf{Entity Changing}, changing entities during the image generation process. For more results and the workflow of these processes, please refer to \textcolor{red}{Supplementary}.
Both quantitative and qualitative evaluations demonstrate the superior performance of our method, even without iterative user feedback. Crucially, our framework inherently further supports multi-round image refinement, a vital capability for real-world instant preference alignment. This includes real-time adjustments to extracted keywords, prompt content, and layout planning. Fig.~\ref{fig:interactive_caption} showcases detailed examples of real-user interactions with our system, illustrating how generated images achieve full alignment with user preferences. Refer to \textcolor{black}{Supp} for more results.

% \begin{figure}
%     \centering
%     \includegraphics[width=0.95\linewidth]{pics/full_interactive_crop.pdf}
%     \caption{Multi-Round Image Updating. Our workflow supports adjusting preferences or updating content to better align with individual user preferences. For entity changes, the modified entities are highlighted with red rectangles.}
%     \label{fig:full_interactive}
% \end{figure}

\section{Conclusion}
\label{sec:conclusion}
% We propose an instant preference-aligned T2I framework capable of understanding and generating images that align with dynamic and multi-dimensional preferences. These preferences are derived from a user-preferred reference image, a base prompt, and real-time user feedback. We comprehensively analyze preference understanding and develop effective strategies to guide existing diffusion models based on these insights. Extensive experiments demonstrate that our approach outperforms previous preference learning and customized generation methods while enabling user-friendly multi-round interactions. 

This paper introduces a training-free framework for instant preference-aligned T2I generation by decoupling preference understanding and guidance. Leveraging MLLM priors, our method enables real-time, fine-grained control over diffusion models without additional tuning. Experiments show superior performance and support for interactive multi-turn refinement, paving the way for preference-aware and dialog-based image generation.

\noindent \textbf{Limitations and Future Work. } For preference understanding, due to the limitations of training data of MLLMs, certain minor preferences may not be fully captured (Table~\ref{tab:keywords_alignment}, ``Others''). This can be mitigated by incorporating limited human feedback (Fig.~\ref{fig:interactive_caption}). For generations, current T2I models struggle with certain unseen conditions, as shown in failure cases in \textcolor{black}{Supp.~C}. 
% ？？？？？？？？？

% \clearpage
\appendix
\setcounter{secnumdepth}{2}
\setcounter{page}{1}
\setcounter{section}{0}

\renewcommand\thesection{\Alph{section}}
\renewcommand{\theequation}{S\arabic{equation}}
\renewcommand{\thefigure}{S\arabic{figure}}
\renewcommand{\thetable}{S\arabic{table}}

% \twocolumn[{
% \renewcommand\twocolumn[1][]{#1}
% \maketitlesupplementary
% % \vspace{-0.5cm}
% % \begin{center}
% %     \textcolor{Red}{\textbf{Warning:} This material may contain disturbing, distressing, offensive, or uncomfortable content.}
% % \end{center}
% }]

This supplementary material provides additional details as follows.

\begin{enumerate}
\setlength{\itemsep}{4pt}
    \item[\ref{supsec:implementation_details}.] Implementation Details.
    \item[\ref{supsec:more_comparison}.] More Comparison.
    \item[\ref{supsec:more_analysis}.] More Analysis.
    \item[\ref{supsec:application}.] Application.
    
\end{enumerate}

\newpage

\section{Implementation Details}
\label{supsec:implementation_details}

\subsection{Datasets}
\begin{figure*}[!th]
    \centering
    \includegraphics[width=0.9\linewidth]{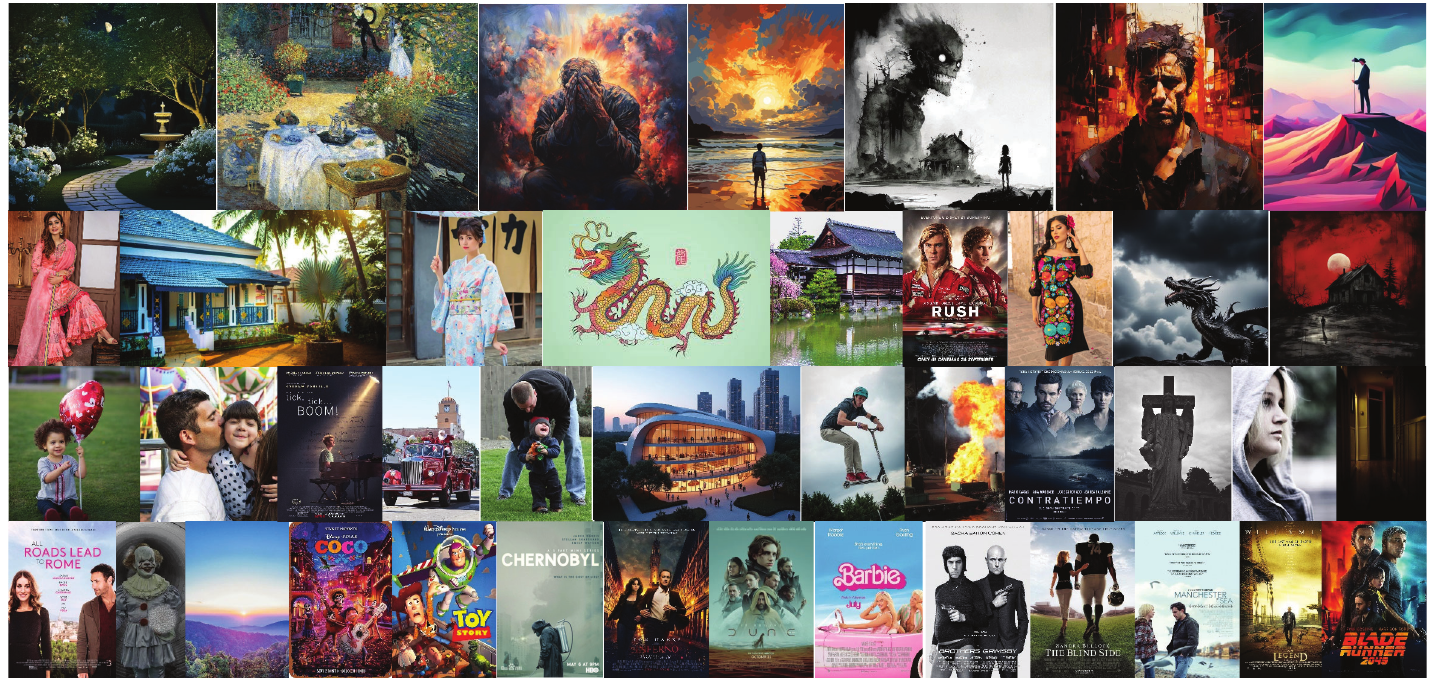}
    \vspace{-0.2cm}
    \caption{An illustration of the query photos. The dataset covers four categories of images: art, culture, emotion, and movie.}
    \label{supfig:dataset_show}
    \vspace{-0.2cm}
\end{figure*}

We collect image data from four categories: culture, art, emotion, and movie data. Emotional pictures are selected from EmoSet~\cite{yang2023emoset}, while the other images are downloaded from the Internet. We show the images of our dataset in Fig.~\ref{supfig:dataset_show}.

\subsection{Prompt Templates for MLLMs}
We employ MLLMs to mimic human reasoning in achieving \textbf{Multi-dimensional Keywords Extraction}, \textbf{Keyword-Conditioned Prompt Enrichment}, and \textbf{Regional Planning}. We show the templates of these modules in the following:
\begin{enumerate}
    \item \textbf{Multi-dimensional Keywords Extraction}:
    
    \begin{tcolorbox}[colback=gray!5!white,colframe=gray!75!black,boxsep=2pt,left=2pt,right=2pt,top=2pt,bottom=2pt]
\textit{Your role is to accurately identify and summarize the artistic style of images, focusing on the art movements and visual techniques. You are familiar with a wide range of traditional and modern art movements (such as Expressionism, Surrealism, Cubism, etc.) and visual techniques (like oil painting, watercolor, digital painting, sculpture, etc.). Your task is to carefully examine images, identify relevant art movements and visual techniques, and generate a list of keywords that succinctly capture the artistic style. You should rely on both your deep understanding of art history and your ability to analyze the visual elements of the image to provide an informed and precise response as follows: 1.Begin by identifying whether the image aligns with any known art movements (e.g., Expressionism, Cubism, Abstract Expressionism, Surrealism, etc.). Look for visual cues such as emotional expression, color contrast, abstraction, or symbolic elements that may indicate a specific movement. 2.focus on the techniques used to create the image. Check for traditional techniques (like oil painting, fresco, or watercolor) or modern ones (such as digital painting, mixed media, or graffiti). 3.Based on the art movements and visual techniques identified, provide 5 summary keywords. If the image corresponds to one of the listed art movements or techniques (e.g., Expressionism, Oil Painting), include those. If there are no direct matches, use your knowledge to suggest relevant keywords based on the visual characteristics you observed. If you are unable to identify enough matches to reach 5 keywords, provide as many relevant keywords as possible, even if the total is fewer than 5. The output should follow the format of the examples below:
Example 1 (Van Gogh's Starry Night):
Keywords: Expressionism, Oil Painting, Impasto, Post-Impressionism
Example 2 (Digital Concept Art):
Keywords: Surrealism, Digital Painting, Neon Sculpture, Cyberpunk Aesthetic.}
\end{tcolorbox}
\item \textbf{Keyword-Conditioned Prompt Enrichment}:
\begin{tcolorbox}[colback=gray!5!white,colframe=gray!75!black,boxsep=2pt,left=2pt,right=2pt,top=2pt,bottom=2pt]
\textit{You are a creative conceptual artist skilled in multimodal narrative design.
You need to expand the simple prompt based on these preference categories of keywords: Artistic Style, Emotional/Atmospheric Preferences, Thematic Preferences, Visual Elements preferences, and other Preferences and the preference signals expressed in the provided image.
Please rearrange the simple prompt as follows: 1.Identify the main objects or key elements and their attributes in the simple input prompt that you will focus on expanding. Note any specific objects, emotions, or concepts that are already part of the original prompt. 2.Based on the preferences given (Artistic Style, Emotional/Atmospheric, Thematic, Visual Elements, and Others), determine 2-4 objects or elements that should be added to the scene to better align the final image with the given preferences. You can add objects that reflect the theme, mood, and other preferences described. For Example: (1)Emotional/Atmospheric: Tension and Gloom suggest solitary figures, dark ocean, foggy landscapes, or dramatic weather conditions. (2)Thematic: Urbanization suggests incorporating elements like decaying buildings, roads, and vehicles.
(3)Others: If there are any specific objects (like a special cat), integrate them into the scene description, ensuring they are contextually and visually aligned with the other preferences. 3. For each object, whether from the original prompt or newly added, provide a detailed objective description based on the preferences. Keep each description under 30 words. 4. Merge the original simple prompt with the objects and their descriptions to form a more complex, concise prompt that reflects all the preferences. Ensure that the added elements naturally flow and integrate with the original context and preferences (40 words at most). And create a simple background prompt (10 words) that encapsulates the preferences but does not include any specific objects mentioned previously.}
\end{tcolorbox}
\begin{tcolorbox}
[colback=gray!5!white,colframe=gray!75!black,boxsep=2pt,left=2pt,right=2pt,top=2pt,bottom=2pt]
\textit{
 The output should follow the format of the examples below:}
\end{tcolorbox}
\item \textbf{Regional Planning}:
\begin{tcolorbox}[colback=gray!5!white,colframe=gray!75!black,boxsep=2pt,left=2pt,right=2pt,top=2pt,bottom=2pt]
\textit{You are an expert in image composition, skilled in interpreting complex scene descriptions and creating spatial arrangements based on detailed prompts and preference keywords. Your task is to:
1.Identity the key objects from the detailed object prompt.
2.Assign spatial positions for each object. This layout assignment should strictly follow the rules below. (1) Basic rules: (a). The image coordinates are based on a system where the top-left corner is [0, 0] and the bottom-right corner is [1, 1]. (b). Assign each object a rectangular space in the image, represented in the format: [top-left x, top-left y, bottom-right x, bottom-right y]. (c). Each object should be assigned a distinct space that doesn't overlap with other objects. (d). Use the visual elements preference keywords that relate to the layout (e.g., Asymmetry, Geometric shapes, High contrast, rule of thirds) to determine how to position the objects within the image. These keywords will guide your decision on object placement in terms of layout.
(2) Layout rules for each object: (a). When assigning spaces to each object, place them in a logical, left-to-right, top-to-bottom manner. (b). No object should exceed the boundaries of the image (i.e., all coordinates should stay within [0, 0] to [1, 1]). (c). Each area should be dedicated to a single object.
3. Finally, in each detailed object prompt, add the location description of that specific object into the original prompt(e.g., on the top, in the left, stay in the center). The output should follow the format of the examples below:}
\end{tcolorbox}
\end{enumerate}
We utilize Chain-of-Thought (CoT) prompting to guide MLLMs in reasoning step by step, enabling them to effectively complete the corresponding task with the assistance of in-context examples, much like how humans approach similar tasks. 
\subsection{Prompt Templates for GPT-4o as Evaluator}
We utilize GPT-4o~\cite{hurst2024gpt} as a human-like evaluator, instructing it to analyze the preference alignment of the generated images from our method and other approaches. The process leverages the few-shot in-context preference learning capability of MLLMs, as demonstrated in prior work~\cite{lee2024prometheus, yu2024few}. The prompt used for this evaluation is provided below. 
\begin{tcolorbox}
[colback=gray!5!white,colframe=gray!75!black,boxsep=2pt,left=2pt,right=2pt,top=2pt,bottom=2pt]
\textit{Part 1: Preference Analysis of a Reference Image
You are an expert in visual aesthetics and human preferences for images. Your task is to analyze the following reference image provided by the user and understand why a person might find it appealing.
Please analyze the image based on the following five aspects: 1.Art Style, 2. Emotion, 3. Theme, 4. Visual Elements, 5. Others (Optional).
After analyzing these five aspects, summarize why you believe a user might like this image overall. Provide a concise explanation that integrates your analysis from the five aspects above.
Here is the reference image:
Part 2: Image Preference Matching from a Set of Images
Then you will be provided with a set of new images. Your task is to choose ONE image from this set that you believe the user would prefer the most, based on the preference profile you established in Part 1 from the reference image. Analyze each image in the set, considering the same five aspects (Art Style, Emotion, Theme, Visual Elements, and Others) you used in Part 1. 
Choose only ONE image from the set that you believe is the closest match to the user's preference.
Format your response as follows:
Chosen Image: Image idx
Explanation of Choice: Image 3 is chosen because...
Here is the set of new images:}
\end{tcolorbox}

\subsection{Definition of each preference category}
\noindent\textbf{Artistic Style:} focuses on the artistic expression and the visual treatment of the image. It includes traditional and modern artistic movements, as well as specific techniques or styles. e.g., Expressionism, Romanticism, Fauvism, Abstract Expressionism, Oil Painting, Watercolor, Acrylics, Tempera, Fresco, Impasto...

\noindent\textbf{Emotional/Atmospheric:} about the mood or feeling that the image should evoke. It can reflect both the emotional tone and the general atmosphere of the image. e.g., Sadness, Joy, Nostalgia, Anger, Fear, Love, Peace, Mystery, Serenity, Tension, Elegance, Melancholy, Romanticism...

\noindent\textbf{Thematic:} encompasses the subjects, themes, and cultural references in the image. It could refer to specific cultural representations, visual narratives, or conceptual themes. e.g., time, Transience, Cycles of Nature, Survival, Decay, Transformation, Greek Mythology, Christianity, Hinduism, Buddhism, Ancient Egypt, Judaism, Norse Mythology, Christian Allegory, Astrological Symbols...

\noindent\textbf{Visual elements:} this category refers to the structural and formal aspects of the image, focusing on the specific visual features such as composition, lines, colors, and textures. e.g., Symmetry, Asymmetry, Rule of thirds, Warm, Cool, Rough, Smooth, Glossy, Straight lines, Curved lines, Organic lines, High contrast, Low contrast...

\section{More Comparison}
\label{supsec:more_comparison}
\subsection{Consistency with the Prompt Intent}
\begin{figure}[th]
    \centering
    \includegraphics[width=0.95\linewidth]{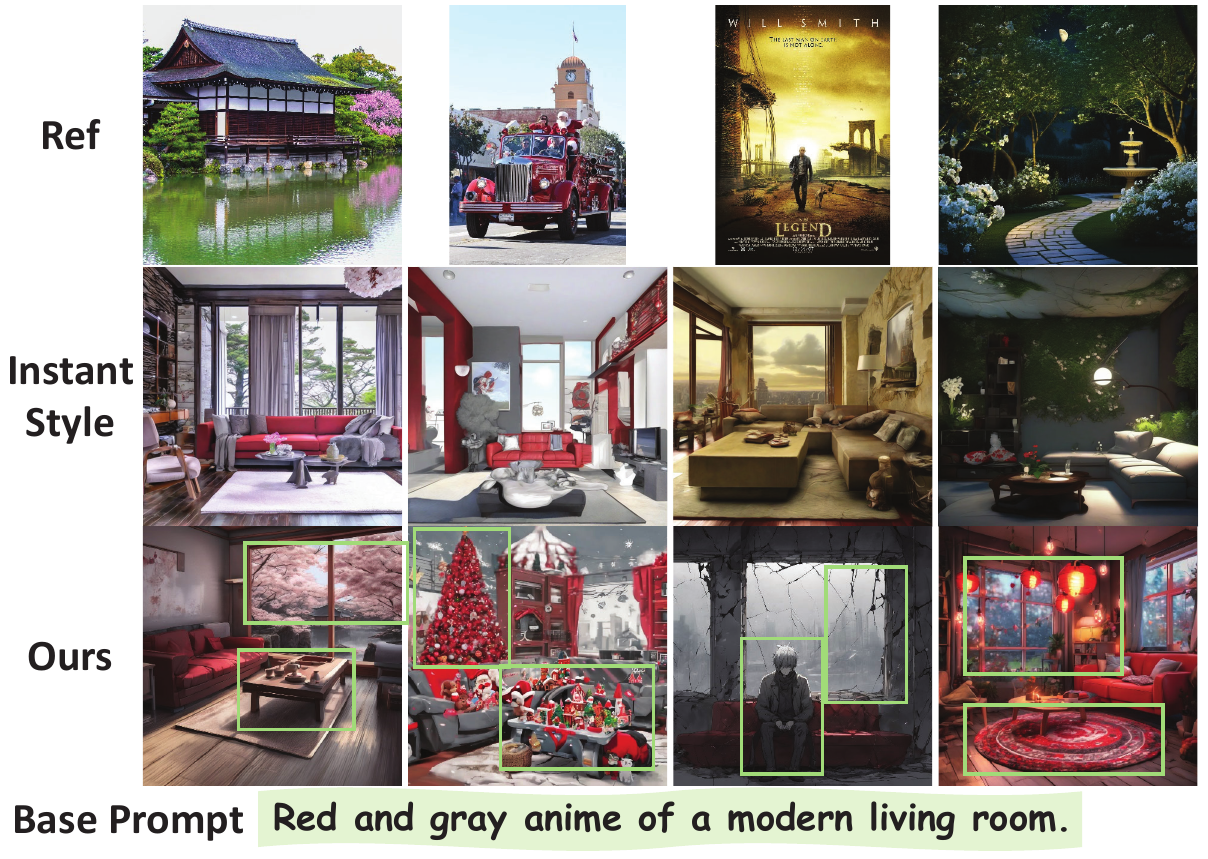}
    \caption{Ablation study of generating images with a specified color palette. Compared with InstantStyle~\cite{wang2024instantstyle} our method can maintain the characteristics of base prompt while achieving preference alignment by incorporating entities.}
    \label{supfig:abl_color}
    \vspace{-0.2cm}
\end{figure}

In our experiments, we observe that the generated images tend to align with a specific color palette. 
To demonstrate that our method ensures coherence between the generated images and the base prompt, we present results in Fig.~\ref{supfig:abl_color}, where prompts with specified color palettes are used. Compared to Style Transfer~\cite{wang2024instantstyle}, our outcomes not only adhere to the base prompt but also incorporate reference-related entities, better aligning with the reference image (e.g., incorporating theme-related objects and conveying a mood consistent with the reference image). 

\subsection{Execution Time $\&$ Alignment Evaluation}
We further compare the execution time and the thematic, visual elements alignment with the other methods. The results are depicted in Tab.~\ref{tab:moremetric}. Our method achieves superior thematic alignment and competitive visual element consistency compared to the training-based InstantStyle, without incurring the substantial time overhead associated with methods like Viper and Fabric.

\begin{table}[!h]
    \centering
    \tiny
    \vspace{-0.3cm}
    \caption{Execution Time \& Alignment Evaluation.}
    \begin{tabular}{cccccccc}
    \toprule
    & NP & Fabric & Mgie & Rewrite & InstantStyle & Viper & Ours \\
    \midrule[0.1pt]
    Thematic$\uparrow$ & 3.67 & 3.45 & 4.53 & 6.03 & 5.12 & 5.32 & \textbf{7.43} \\
    \makecell{Visual\\Elements$\uparrow$} & 3.42 & 3.57 & 4.12 & 5.49 & \textbf{6.83} & 6.05 & 6.32 \\
    \makecell{Execution\\Time (s)} & $\sim$7 & $>$180 & $\sim$12 & $\sim$100 & $\sim$10 & $>$180 & $\sim$180 \\
    \bottomrule
    \end{tabular}
    % \vspace{-0.3cm}
    \label{tab:moremetric}
\end{table}

\subsection{Comparing Our Global Preference Guidance with Viper}
\begin{figure*}[!th]
    \centering
    \includegraphics[width=0.95\linewidth]{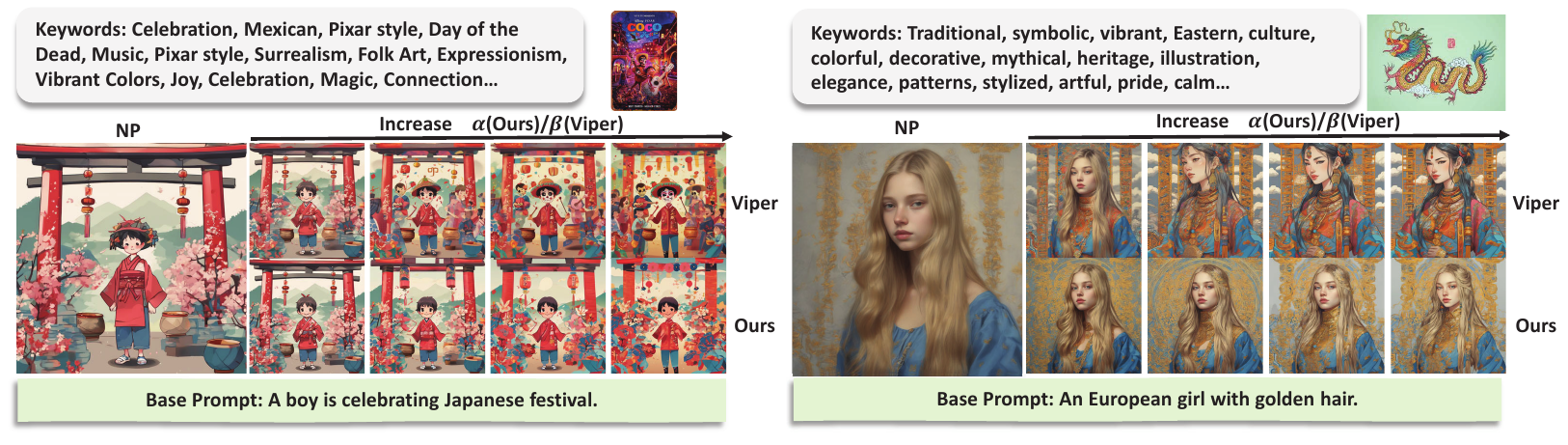}
    \caption{We further give a detailed comparison of our \textbf{Global Preference Guidance} with the preference-aligned generation method Viper~\cite{salehi2024viper}. }
    \label{supfig:comp_viper}
\end{figure*}

% viper也无法用到最新的如flux-dev等模型上
The main comparison results in the main paper demonstrate that Viper~\cite{salehi2024viper} struggles to generate images that align with the extracted keywords. This limitation can be attributed to: (1) Missing Preference-Aligned Entities, which our proposed prompt enrichment module and local cross-attention modulation effectively address; and (2) The Trade-off Between Preference Alignment and Prompt Fidelity, where their approach tends to distort the original prompt, particularly when conflicts arise with preference keywords. 
Formally, the generation process of Viper can be formulated as:
\begin{align}
    \epsilon_{vp}(\bm{x}_{t} t)&=\epsilon(\bm{x}_{t}, y_{\text{key}}, t) \\
    \epsilon_\theta(\bm{x}, t, y)&=(1-\omega)\epsilon(\bm{x}, t)+\omega(\epsilon_\theta(\bm{x}_{t}, t, y)+\beta\epsilon_{vp})
\end{align}
where $\epsilon$ is the image generation model and $\omega$ is the guidance scale. Viper~\cite{salehi2024viper} operates within the latent space and relies on classifier-free guidance. This design fundamentally restricts its generalization, particularly to models like Flux.1-dev, and risks entangling semantic disturbances that significantly alter the original semantics. In contrast, our \textbf{Global Preference Guidance}:
\begin{equation}
    \bm{P}_{y_{i}} = \mathcal{T}_{text}(y_{i}) + \alpha \cdot\text{Proj}_{\perp}(\mathcal{T}_{text}(y_{\text{key}})),
\end{equation}
is deployed in the text embedding space. It leverages a projection operation to subtract content related to the current semantic context, thereby ensuring semantic preservation during preference alignment.
In Fig.~\ref{supfig:comp_viper}, we assess the performance of our \textbf{Global Preference Guidance} and Viper~\cite{salehi2024viper} in original context preserving. Viper~\cite{salehi2024viper} struggles to maintain contextual integrity while adhering to preference signals derived from keywords (e.g., adding a hat to the Japanese boy and turning him into a Mexican boy or generating a girl with an Eastern appearance without golden hair on the right side of the figure). Our \textbf{Global Preference Guidance} fully preserves the original context while making essential modifications to align with the preference keywords.

\subsection{Comparing Our Local Cross Attention Modulation with RPG}
\begin{figure*}[!th]
    \centering
    \includegraphics[width=0.95\linewidth]{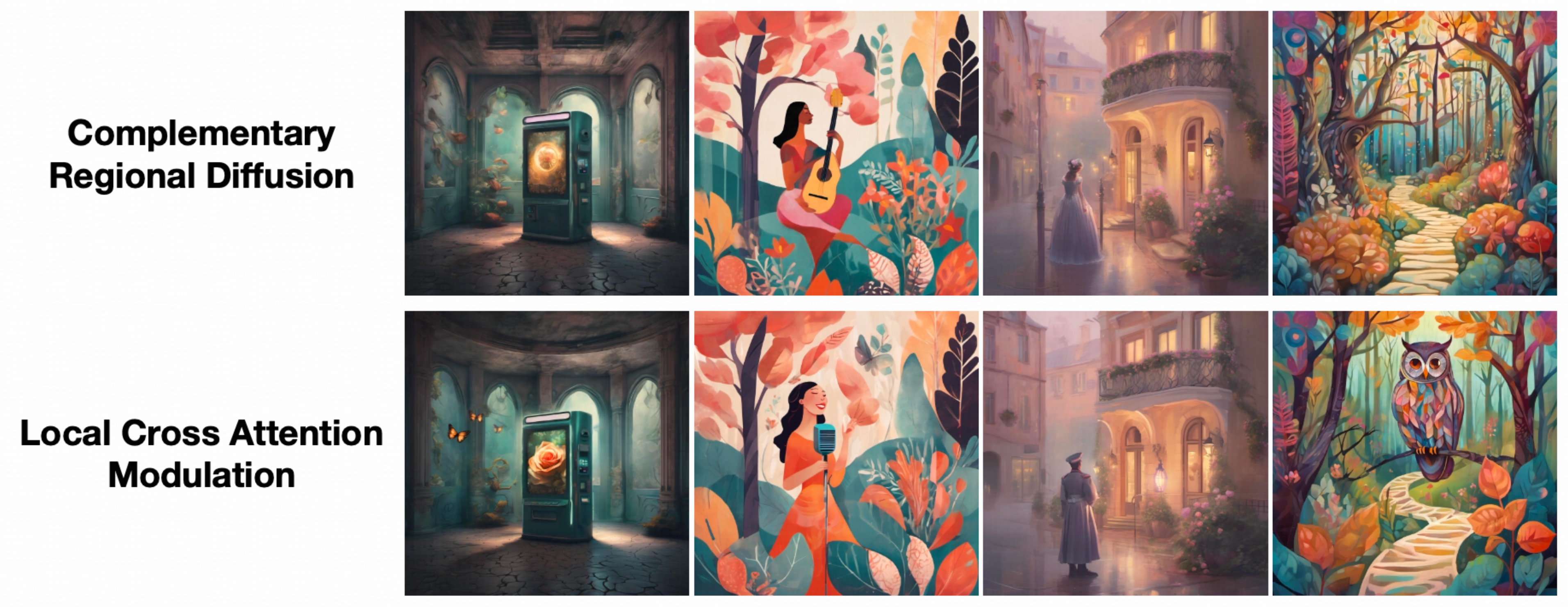}
    \caption{We further give a detailed qualitative comparison of our \textbf{Local Cross Attention Modulation} with the \textbf{Complementary Regional Diffusion} used in RPG. The RPG method can lead to semantic confusion among entities, resulting in poor alignment where preference-aligned entities are often omitted. The prompts are: (1) ``Old vending machine in school hallway'', (2) ``A woman singing into a microphone'', (3) ``A man in military uniform'', (4) ``A barred owl peeking out from tree branches''. }
    \label{supfig:comp_rpg}
\end{figure*}

RPG~\cite{yang2024mastering} employs a coarse Complementary Regional Diffusion for compositional generation and utilizes resizing on the latent image feature to ensure each entity's appearance. However, this coarse compositional control often leads to unsatisfactory image generation for scenes containing multiple entities with intricate positions and action relationships. Furthermore, the resizing operation itself disrupts the latent space, resulting in poor performance characterized by blurring effects, as evidenced in Tab.~\ref{tab:comprpg} and Fig.~\ref{supfig:comp_rpg}.
In Tab.~\ref{tab:comprpg}, we compare the compositional generation performance between our methods and that of RPG. Our method enhances object arrangement and yields superior compositional generation.
\begin{table}[]
    \centering
    \vspace{-0.4cm}
    \captionof{table}{Comparison on T2I-CompBench.}
    \begin{tabular}{ccc}
    \toprule
        Method & Spatial$\uparrow$ & Non-Spatial$\uparrow$ \\
    \midrule[0.1pt]
        RPG & 0.468 & 0.352 \\
        Ours & \textbf{0.603} & \textbf{0.528} \\
    \bottomrule
    \end{tabular}
    % \vspace{-0.3cm}
    \vspace{-0.4cm}
    \label{tab:comprpg}
\end{table}

\section{More Analysis}
\label{supsec:more_analysis}
\subsection{Ablation on Global Preference Guidance}
\begin{figure*}[!th]
    \centering
    \includegraphics[width=0.95\linewidth]{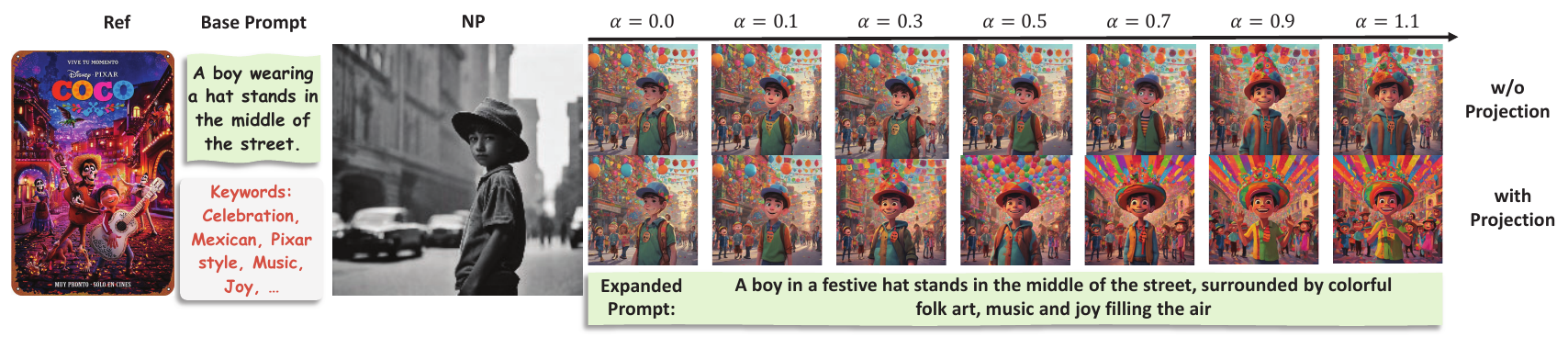}
    \caption{Ablation study of the hyperparameter $\alpha$ and the projection function used in our method.}
    \label{supfig:abl_alpha_proj}
\end{figure*}

\noindent\textbf{Ablation of hyperparameter $\alpha$.} We analyze the effect of $\alpha$ by varying its value as a hyperparameter. The results, shown in Fig.~\ref{supfig:abl_alpha_proj}, demonstrate that as $\alpha$ increases from $0$ to $1.1$, the generated images become more aligned with the reference image, deviating from general outputs while preserving the content specified by the base prompt. 

\noindent\textbf{Ablation of the preference projection.} The project operation ensures that semantic modifications prioritize preference alignment while preserving the context of the original prompt. As shown in Fig.~\ref{supfig:abl_alpha_proj}, without projection, the generated results become less effective and risk disturbing prompt information.

\subsection{Ablation of $\lambda$}
\begin{figure*}
    \centering
    \includegraphics[width=0.95\linewidth]{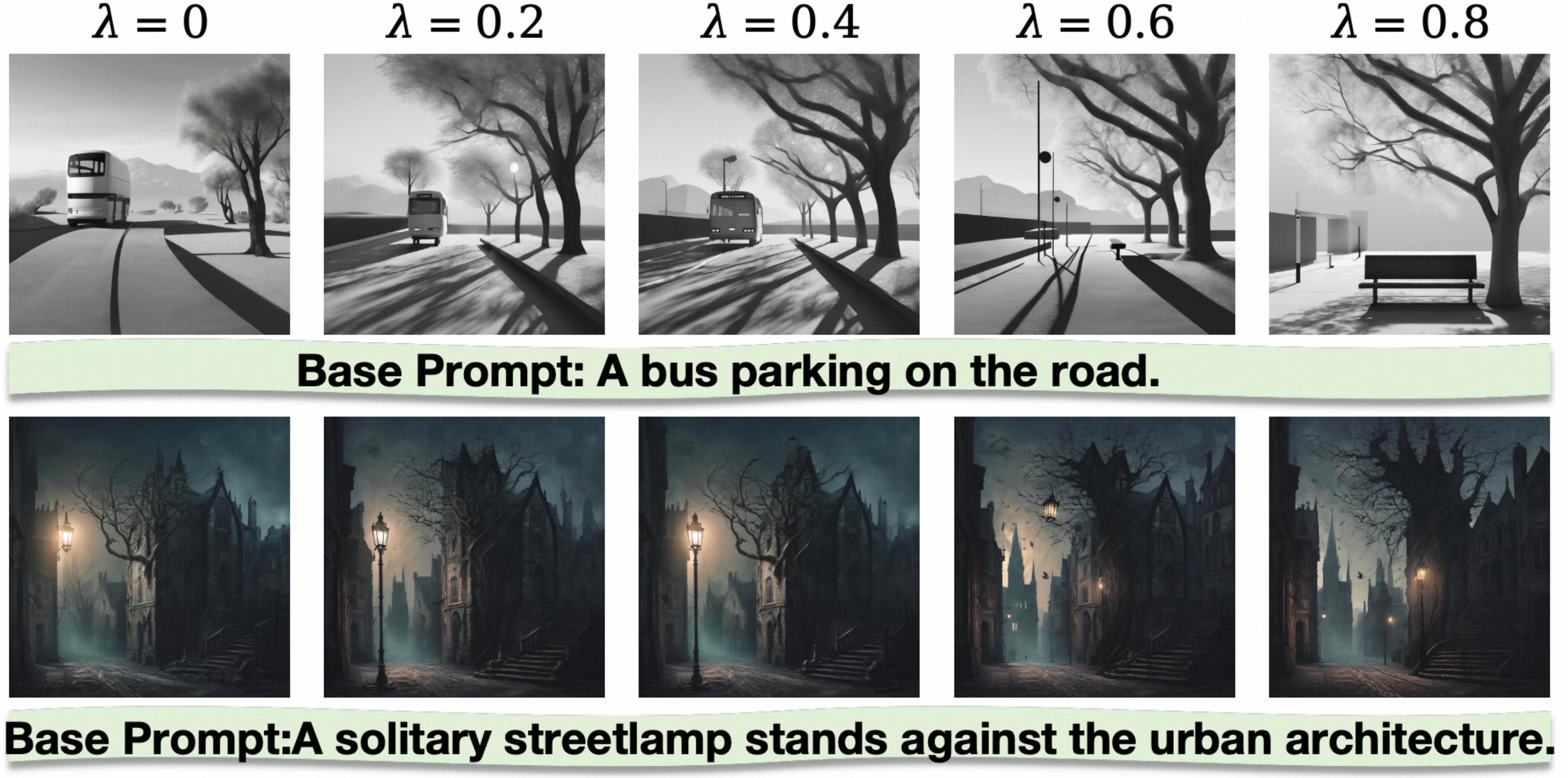}
    \caption{Ablation of the hyperparameter $\lambda$ in \textbf{Local Cross Attention Modulation}.}
    \label{supfig:abl_lambda}
\end{figure*}

% include the ablation study of lambda
In Fig.~\ref{supfig:abl_lambda}, we provide the ablation study of hyperparameter $\lambda$. We find that the proper value of $\lambda$ can benefit the conjunction of each entity, while excessive $\lambda$ could lead to undesirable results.

\subsection{Ablation of Full Pipeline}
\begin{figure*}[!th]
    \centering
    \includegraphics[width=0.95\linewidth]{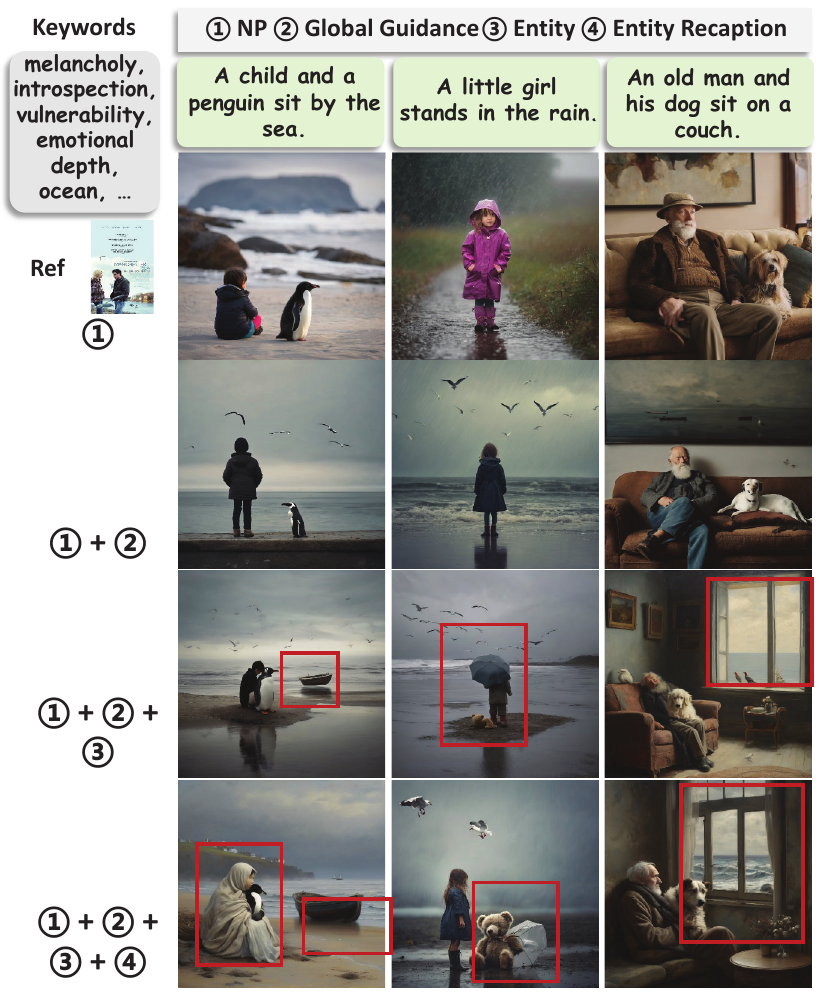}
    \caption{Ablation study of our full pipeline. Global Preference Guidance guides the generated images with preference-aligned elements such as color and atmosphere while preserving the base prompt. By incorporating entities (circled out by the red rectangles), we can further enhance the emotional resonance of the images. Additionally, the recaptioning process provides more descriptive and informative details for the entities, which further improves alignment with preference signals. For example, as we make the boy wearing a blanket in the first column, it enhances a sense of drifting and makes the "vulnerability" more prominent.}
    \label{supfig:abl_textexp}
\end{figure*}

We conduct an ablation study of our full pipeline to evaluate the effect of each module. The results are presented in Fig~\ref{supfig:abl_textexp}. With \textbf{Global Preference Guidance}, the generated images tend to preserve the base prompt while modifying elements such as color and atmosphere. However, this approach alone struggles to align with thematic-related preferences, such as ``ocean'' that leads to ``deep thinking''. By incorporating ``ocean'' and other entities circled out in red boxes, we can further enhance the emotional resonance of the images, which is achieved by the \textbf{Prompt Enrichment} and \textbf{Local Cross-attention Modulation}. Additionally, the recaptioning process provides more descriptive and informative details for the entities, which further improves alignment with preference signals. For example, by having the boy wrapped in a blanket in the first column, we enhance the sense of drifting and makes the feeling of "vulnerability" more prominent.

\section{Application}
\label{supsec:application}
\subsection{Multi-Round Image Updating}
\begin{figure*}
    \centering
    \includegraphics[width=0.95\linewidth]{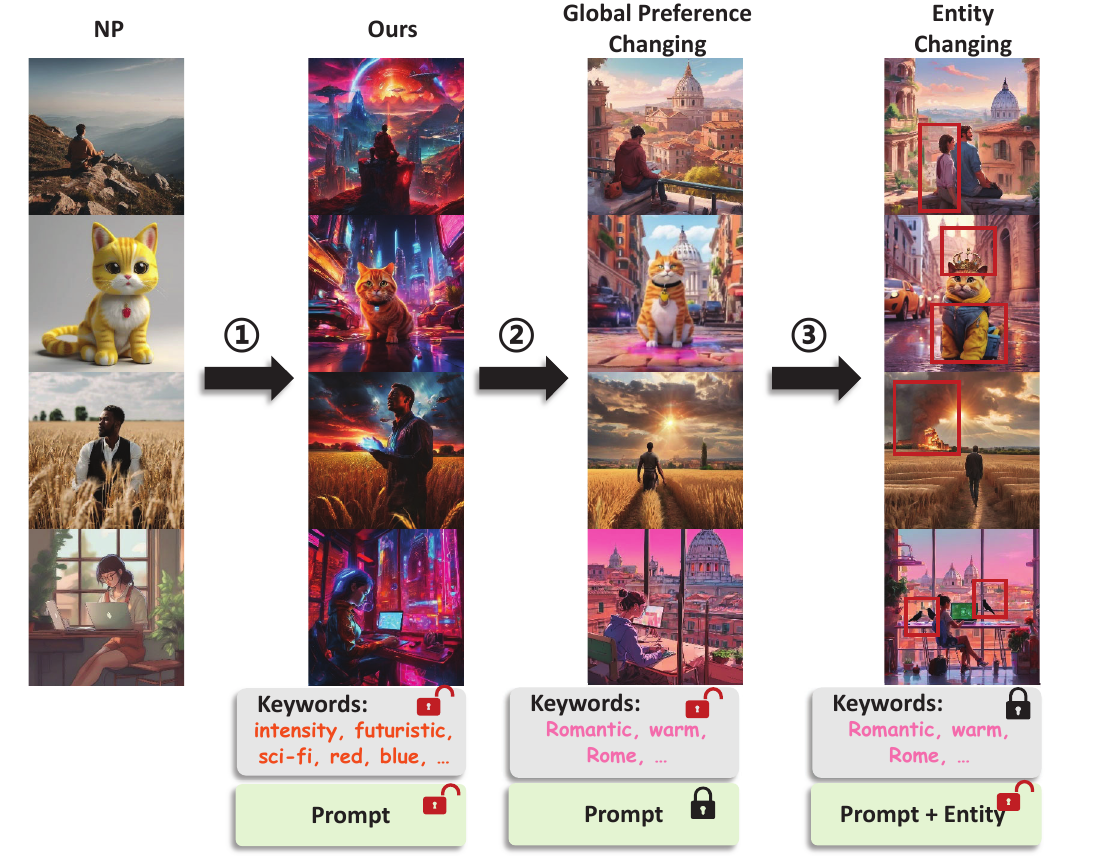}
    \caption{Multi-Round Image Updating. Our framework supports adjusting preferences or updating content to better align with individual user preferences. For entity changes, the modified entities are highlighted with red rectangles.}
    \label{supfig:full_interactive}
    % \vspace{-0.3cm}
\end{figure*}

Here, we present additional results of Multi-Round Image Updating, as depicted in Fig.~\ref{supfig:full_interactive}. Since preferences suggested by MLLMs may not fully align with personal requirements and personal preferences can vary, our franework allows for iterative adjustments and content updating across multiple rounds to better align with individual user preferences. For instance, in step \textbf{\textcircled{2}}, the overall preference shifts from "intensity, futuristic, sci-fi, ..." to "romantic, warm, Rome, ...". In step \textbf{\textcircled{3}}, in the third row, adding a fire reintroduces a sense of "intensity" to the image.

\subsection{Generalizing to Various Backbones}
\begin{figure*}[!ht]
    \centering
    \includegraphics[width=1.0\linewidth]{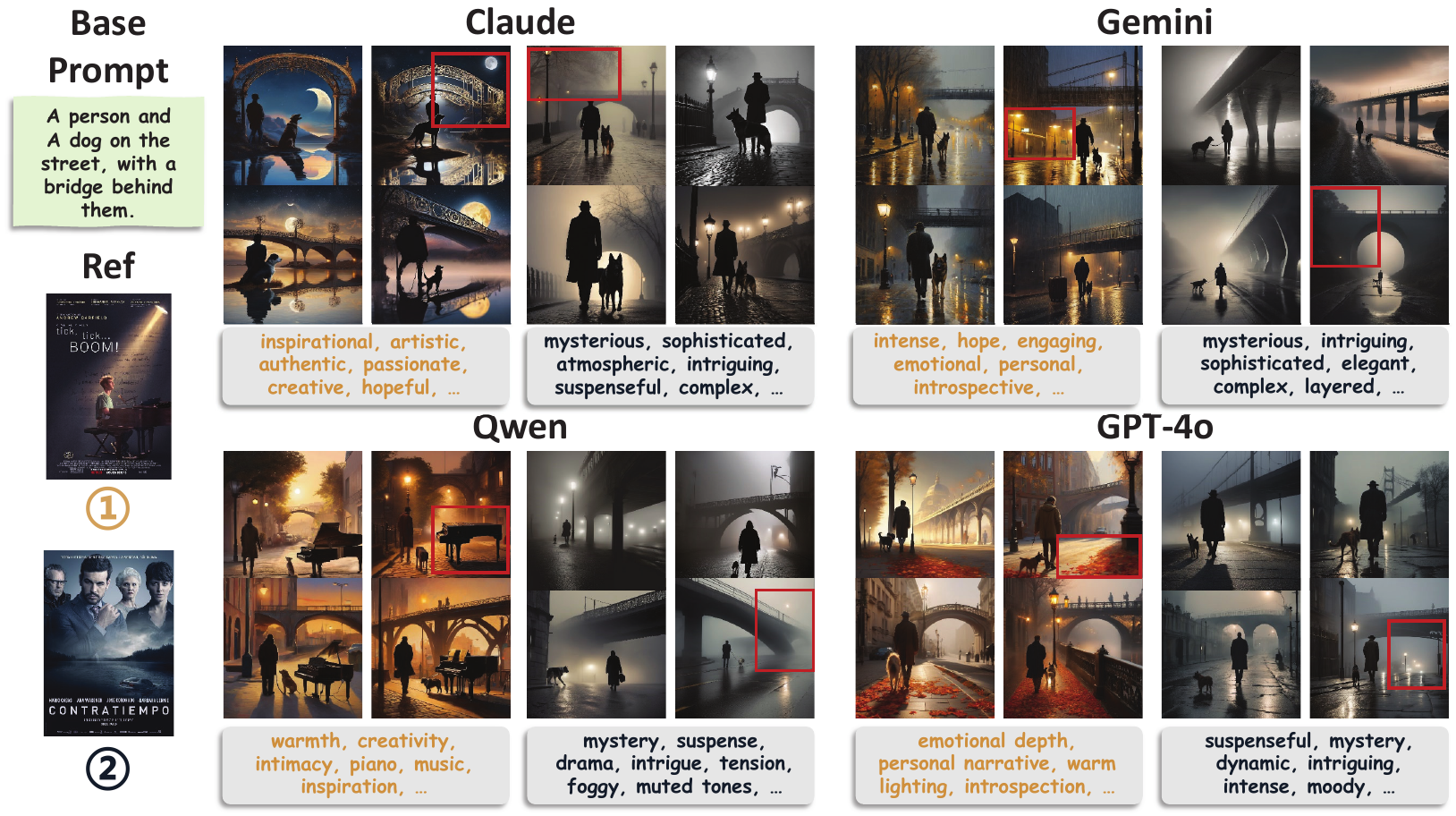}
    \caption{Generalizing our framework to different MLLMs, including Claude~\cite{claude}, Genimi~\cite{team2023gemini}, Qwen~\cite{bai2025qwen2}, and GPT-4o~\cite{hurst2024gpt}.}
    \label{supfig:other_mllms}
\end{figure*}

\begin{figure*}[!th]
    \centering
    \includegraphics[width=0.95\linewidth]{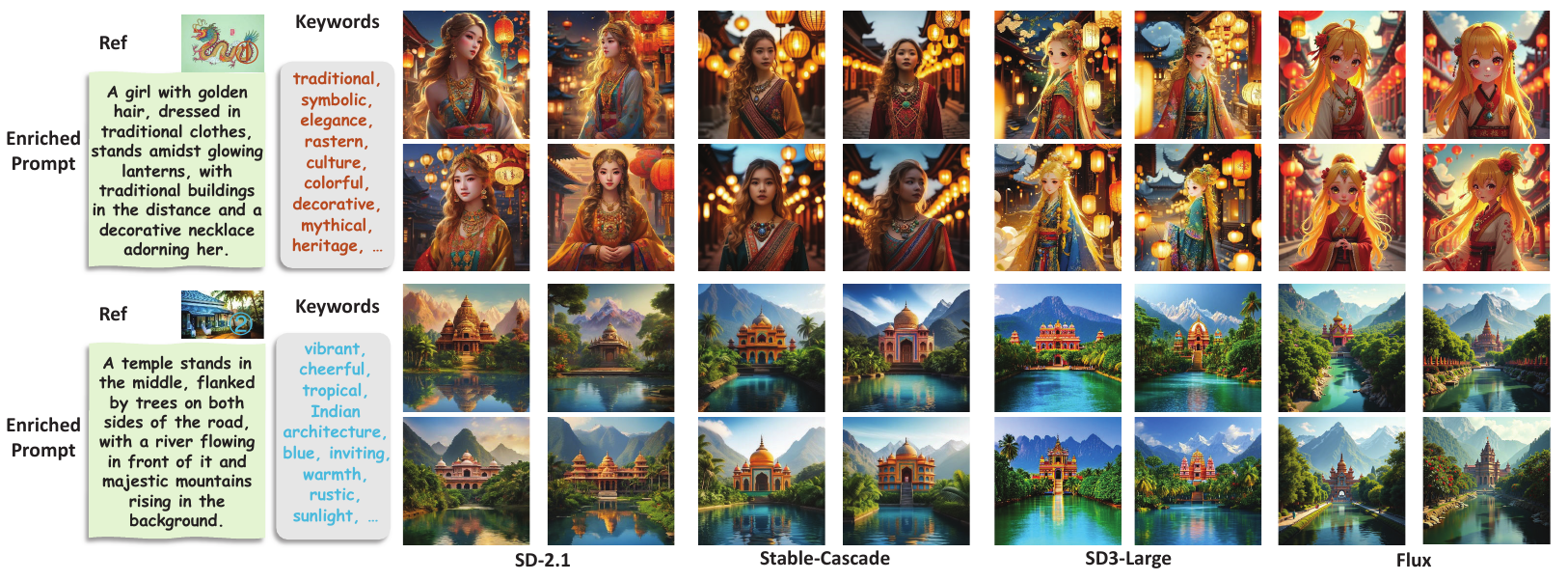}
    \caption{Generalizing our framework to different Diffusion backbones, including SD-2.1~\cite{rombach2022high}, Stable-Cascade~\cite{pernias2023wurstchen}, and SD3-Large~\cite{esser2024scaling}, and Flux.1-dev~\cite{blackforestlabs2024flux}.}
    \vspace{-0.3cm}
    \label{supfig:other_sds}
\end{figure*}

Our framework exhibits a high generalization ability, capable of generalizing to a large range of MLLMs (Fig.~\ref{supfig:other_mllms}) and Stable Diffusion backbones (Fig.~\ref{supfig:other_sds}). 

\noindent\textbf{(1) Generalizing to Other MLLMs: }As depicted in Fig.~\ref{supfig:other_mllms}, the preference signals extracted by different MLLMs may vary to some extent, for example, in the case of the first reference image. However, the extracted preferences remain closely aligned with the reference images. Additionally, the flexibility of our framework supports preference adjustments, as demonstrated in the previous \textbf{Sec. }\textbf{Multi-Round Image Updating}.

\noindent\textbf{(2) Generalizing to Other Diffusion Models: } As depicted in Fig.~\ref{supfig:other_sds}, the choice of diffusion backbones can influence the generation results; however, the outputs remain consistent with the prompt and preference keywords.

\bibliography{AnonymousSubmission/LaTeX/aaai2026}

\begin{thebibliography}{51}
\providecommand{\natexlab}[1]{#1}

\bibitem[{Anthropic(2024)}]{claude}
Anthropic. 2024.
\newblock The Claude 3 Model Family: Opus, Sonnet, Haiku.

\bibitem[{Avrahami et~al.(2023)Avrahami, Hayes, Gafni, Gupta, Taigman, Parikh, Lischinski, Fried, and Yin}]{avrahami2023spatext}
Avrahami, O.; Hayes, T.; Gafni, O.; Gupta, S.; Taigman, Y.; Parikh, D.; Lischinski, D.; Fried, O.; and Yin, X. 2023.
\newblock Spatext: Spatio-textual representation for controllable image generation.
\newblock In \emph{Proceedings of the IEEE/CVF Conference on Computer Vision and Pattern Recognition}, 18370--18380.

\bibitem[{Bai et~al.(2025)Bai, Chen, Liu, Wang, Ge, Song, Dang, Wang, Wang, Tang et~al.}]{bai2025qwen2}
Bai, S.; Chen, K.; Liu, X.; Wang, J.; Ge, W.; Song, S.; Dang, K.; Wang, P.; Wang, S.; Tang, J.; et~al. 2025.
\newblock Qwen2. 5-VL Technical Report.
\newblock \emph{arXiv preprint arXiv:2502.13923}.

\bibitem[{Black et~al.(2023)Black, Janner, Du, Kostrikov, and Levine}]{black2023training}
Black, K.; Janner, M.; Du, Y.; Kostrikov, I.; and Levine, S. 2023.
\newblock Training diffusion models with reinforcement learning.
\newblock \emph{arXiv preprint arXiv:2305.13301}.

\bibitem[{Chatterjee et~al.(2025)Chatterjee, Stan, Aflalo, Paul, Ghosh, Gokhale, Schmidt, Hajishirzi, Lal, Baral et~al.}]{chatterjee2025getting}
Chatterjee, A.; Stan, G. B.~M.; Aflalo, E.; Paul, S.; Ghosh, D.; Gokhale, T.; Schmidt, L.; Hajishirzi, H.; Lal, V.; Baral, C.; et~al. 2025.
\newblock Getting it right: Improving spatial consistency in text-to-image models.
\newblock In \emph{European Conference on Computer Vision}, 204--222. Springer.

\bibitem[{Chefer et~al.(2023)Chefer, Alaluf, Vinker, Wolf, and Cohen-Or}]{chefer2023attend}
Chefer, H.; Alaluf, Y.; Vinker, Y.; Wolf, L.; and Cohen-Or, D. 2023.
\newblock Attend-and-excite: Attention-based semantic guidance for text-to-image diffusion models.
\newblock \emph{ACM Transactions on Graphics (TOG)}, 42(4): 1--10.

\bibitem[{Chen, Laina, and Vedaldi(2024)}]{chen2024training}
Chen, M.; Laina, I.; and Vedaldi, A. 2024.
\newblock Training-free layout control with cross-attention guidance.
\newblock In \emph{Proceedings of the IEEE/CVF Winter Conference on Applications of Computer Vision}, 5343--5353.

\bibitem[{Chen et~al.(2024)Chen, Zhang, Weng, Pan, and Lan}]{chen2024tailored}
Chen, Z.; Zhang, L.; Weng, F.; Pan, L.; and Lan, Z. 2024.
\newblock Tailored visions: Enhancing text-to-image generation with personalized prompt rewriting.
\newblock In \emph{Proceedings of the IEEE/CVF Conference on Computer Vision and Pattern Recognition}, 7727--7736.

\bibitem[{Clark et~al.(2023)Clark, Vicol, Swersky, and Fleet}]{clark2023directly}
Clark, K.; Vicol, P.; Swersky, K.; and Fleet, D.~J. 2023.
\newblock Directly Fine-Tuning Diffusion Models on Differentiable Rewards.
\newblock arXiv:2309.17400.

\bibitem[{Esser et~al.(2024)Esser, Kulal, Blattmann, Entezari, M{\"u}ller, Saini, Levi, Lorenz, Sauer, Boesel et~al.}]{esser2024scaling}
Esser, P.; Kulal, S.; Blattmann, A.; Entezari, R.; M{\"u}ller, J.; Saini, H.; Levi, Y.; Lorenz, D.; Sauer, A.; Boesel, F.; et~al. 2024.
\newblock Scaling rectified flow transformers for high-resolution image synthesis.
\newblock In \emph{Forty-first International Conference on Machine Learning}.

\bibitem[{Fan et~al.(2024)Fan, Watkins, Du, Liu, Ryu, Boutilier, Abbeel, Ghavamzadeh, Lee, and Lee}]{fan2024reinforcement}
Fan, Y.; Watkins, O.; Du, Y.; Liu, H.; Ryu, M.; Boutilier, C.; Abbeel, P.; Ghavamzadeh, M.; Lee, K.; and Lee, K. 2024.
\newblock Reinforcement learning for fine-tuning text-to-image diffusion models.
\newblock \emph{Advances in Neural Information Processing Systems}, 36.

\bibitem[{Frenkel et~al.(2025)Frenkel, Vinker, Shamir, and Cohen-Or}]{frenkel2025implicit}
Frenkel, Y.; Vinker, Y.; Shamir, A.; and Cohen-Or, D. 2025.
\newblock Implicit style-content separation using b-lora.
\newblock In \emph{European Conference on Computer Vision}, 181--198. Springer.

\bibitem[{Fu et~al.(2023)Fu, Hu, Du, Wang, Yang, and Gan}]{fu2023guiding}
Fu, T.-J.; Hu, W.; Du, X.; Wang, W.~Y.; Yang, Y.; and Gan, Z. 2023.
\newblock Guiding instruction-based image editing via multimodal large language models.
\newblock \emph{arXiv preprint arXiv:2309.17102}.

\bibitem[{Gal et~al.(2022)Gal, Alaluf, Atzmon, Patashnik, Bermano, Chechik, and Cohen-Or}]{gal2022image}
Gal, R.; Alaluf, Y.; Atzmon, Y.; Patashnik, O.; Bermano, A.~H.; Chechik, G.; and Cohen-Or, D. 2022.
\newblock An image is worth one word: Personalizing text-to-image generation using textual inversion.
\newblock \emph{arXiv preprint arXiv:2208.01618}.

\bibitem[{Hertz et~al.(2024)Hertz, Voynov, Fruchter, and Cohen-Or}]{hertz2024style}
Hertz, A.; Voynov, A.; Fruchter, S.; and Cohen-Or, D. 2024.
\newblock Style aligned image generation via shared attention.
\newblock In \emph{Proceedings of the IEEE/CVF Conference on Computer Vision and Pattern Recognition}, 4775--4785.

\bibitem[{Ho, Jain, and Abbeel(2020)}]{ho2020denoising}
Ho, J.; Jain, A.; and Abbeel, P. 2020.
\newblock Denoising diffusion probabilistic models.
\newblock \emph{Advances in neural information processing systems}, 33: 6840--6851.

\bibitem[{Hurst et~al.(2024)Hurst, Lerer, Goucher, Perelman, Ramesh, Clark, Ostrow, Welihinda, Hayes, Radford et~al.}]{hurst2024gpt}
Hurst, A.; Lerer, A.; Goucher, A.~P.; Perelman, A.; Ramesh, A.; Clark, A.; Ostrow, A.; Welihinda, A.; Hayes, A.; Radford, A.; et~al. 2024.
\newblock Gpt-4o system card.
\newblock \emph{arXiv preprint arXiv:2410.21276}.

\bibitem[{Kirstain et~al.(2023)Kirstain, Polyak, Singer, Matiana, Penna, and Levy}]{kirstain2023pickapic}
Kirstain, Y.; Polyak, A.; Singer, U.; Matiana, S.; Penna, J.; and Levy, O. 2023.
\newblock Pick-a-Pic: An Open Dataset of User Preferences for Text-to-Image Generation.
\newblock arXiv:2305.01569.

\bibitem[{Labs(2024)}]{blackforestlabs2024flux}
Labs, B.~F. 2024.
\newblock FLUX: Inference Repository.
\newblock \url{https://github.com/black-forest-labs/flux}.
\newblock Accessed: 2024-10-25.

\bibitem[{Lee et~al.(2024)Lee, Kim, Park, Kim, and Seo}]{lee2024prometheus}
Lee, S.; Kim, S.; Park, S.; Kim, G.; and Seo, M. 2024.
\newblock Prometheus-vision: Vision-language model as a judge for fine-grained evaluation.
\newblock In \emph{Findings of the Association for Computational Linguistics ACL 2024}, 11286--11315.

\bibitem[{Li et~al.(2023)Li, Liu, Wu, Mu, Yang, Gao, Li, and Lee}]{li2023gligen}
Li, Y.; Liu, H.; Wu, Q.; Mu, F.; Yang, J.; Gao, J.; Li, C.; and Lee, Y.~J. 2023.
\newblock Gligen: Open-set grounded text-to-image generation.
\newblock In \emph{Proceedings of the IEEE/CVF Conference on Computer Vision and Pattern Recognition}, 22511--22521.

\bibitem[{Li et~al.(2024)Li, Yang, Wang, and Dong}]{li2024beyond}
Li, Y.; Yang, S.; Wang, W.; and Dong, J. 2024.
\newblock Beyond inserting: Learning identity embedding for semantic-fidelity personalized diffusion generation.
\newblock \emph{arXiv preprint arXiv:2402.00631}.

\bibitem[{Lian et~al.(2023)Lian, Li, Yala, and Darrell}]{lian2023llm}
Lian, L.; Li, B.; Yala, A.; and Darrell, T. 2023.
\newblock Llm-grounded diffusion: Enhancing prompt understanding of text-to-image diffusion models with large language models.
\newblock \emph{arXiv preprint arXiv:2305.13655}.

\bibitem[{Liu et~al.(2022)Liu, Li, Du, Torralba, and Tenenbaum}]{liu2022compositional}
Liu, N.; Li, S.; Du, Y.; Torralba, A.; and Tenenbaum, J.~B. 2022.
\newblock Compositional visual generation with composable diffusion models.
\newblock In \emph{European Conference on Computer Vision}, 423--439. Springer.

\bibitem[{Lu et~al.(2022)Lu, Zhou, Bao, Chen, Li, and Zhu}]{lu2022dpm}
Lu, C.; Zhou, Y.; Bao, F.; Chen, J.; Li, C.; and Zhu, J. 2022.
\newblock Dpm-solver: A fast ode solver for diffusion probabilistic model sampling in around 10 steps.
\newblock \emph{Advances in Neural Information Processing Systems}, 35: 5775--5787.

\bibitem[{Mou et~al.(2024)Mou, Wang, Xie, Wu, Zhang, Qi, and Shan}]{mou2024t2i}
Mou, C.; Wang, X.; Xie, L.; Wu, Y.; Zhang, J.; Qi, Z.; and Shan, Y. 2024.
\newblock T2i-adapter: Learning adapters to dig out more controllable ability for text-to-image diffusion models.
\newblock In \emph{Proceedings of the AAAI Conference on Artificial Intelligence}, volume~38, 4296--4304.

\bibitem[{Pernias et~al.(2023)Pernias, Rampas, Richter, Pal, and Aubreville}]{pernias2023wurstchen}
Pernias, P.; Rampas, D.; Richter, M.~L.; Pal, C.~J.; and Aubreville, M. 2023.
\newblock W{\"u}rstchen: An efficient architecture for large-scale text-to-image diffusion models.
\newblock \emph{arXiv preprint arXiv:2306.00637}.

\bibitem[{Podell et~al.(2023)Podell, English, Lacey, Blattmann, Dockhorn, M{\"u}ller, Penna, and Rombach}]{podell2023sdxl}
Podell, D.; English, Z.; Lacey, K.; Blattmann, A.; Dockhorn, T.; M{\"u}ller, J.; Penna, J.; and Rombach, R. 2023.
\newblock Sdxl: Improving latent diffusion models for high-resolution image synthesis.
\newblock \emph{arXiv preprint arXiv:2307.01952}.

\bibitem[{Prabhudesai et~al.(2023)Prabhudesai, Goyal, Pathak, and Fragkiadaki}]{prabhudesai2023aligning}
Prabhudesai, M.; Goyal, A.; Pathak, D.; and Fragkiadaki, K. 2023.
\newblock Aligning Text-to-Image Diffusion Models with Reward Backpropagation.
\newblock arXiv:2310.03739.

\bibitem[{Radford et~al.(2021)Radford, Kim, Hallacy, Ramesh, Goh, Agarwal, Sastry, Askell, Mishkin, Clark et~al.}]{radford2021learning}
Radford, A.; Kim, J.~W.; Hallacy, C.; Ramesh, A.; Goh, G.; Agarwal, S.; Sastry, G.; Askell, A.; Mishkin, P.; Clark, J.; et~al. 2021.
\newblock Learning transferable visual models from natural language supervision.
\newblock In \emph{International conference on machine learning}, 8748--8763. PMLR.

\bibitem[{Razavi, Van~den Oord, and Vinyals(2019)}]{razavi2019generating}
Razavi, A.; Van~den Oord, A.; and Vinyals, O. 2019.
\newblock Generating diverse high-fidelity images with vq-vae-2.
\newblock \emph{Advances in neural information processing systems}, 32.

\bibitem[{Rombach et~al.(2022)Rombach, Blattmann, Lorenz, Esser, and Ommer}]{rombach2022high}
Rombach, R.; Blattmann, A.; Lorenz, D.; Esser, P.; and Ommer, B. 2022.
\newblock High-resolution image synthesis with latent diffusion models.
\newblock In \emph{Proceedings of the IEEE/CVF conference on computer vision and pattern recognition}, 10684--10695.

\bibitem[{Ruiz et~al.(2023)Ruiz, Li, Jampani, Pritch, Rubinstein, and Aberman}]{ruiz2023dreambooth}
Ruiz, N.; Li, Y.; Jampani, V.; Pritch, Y.; Rubinstein, M.; and Aberman, K. 2023.
\newblock Dreambooth: Fine tuning text-to-image diffusion models for subject-driven generation.
\newblock In \emph{Proceedings of the IEEE/CVF conference on computer vision and pattern recognition}, 22500--22510.

\bibitem[{Salehi et~al.(2024)Salehi, Shafiei, Yeo, Bachmann, and Zamir}]{salehi2024viper}
Salehi, S.; Shafiei, M.; Yeo, T.; Bachmann, R.; and Zamir, A. 2024.
\newblock ViPer: Visual Personalization of Generative Models via Individual Preference Learning.
\newblock \emph{arXiv preprint arXiv:2407.17365}.

\bibitem[{Shen et~al.(2024)Shen, Zhang, Zhao, Zhu, and Xiao}]{shen2024pmg}
Shen, X.; Zhang, R.; Zhao, X.; Zhu, J.; and Xiao, X. 2024.
\newblock Pmg: Personalized multimodal generation with large language models.
\newblock In \emph{Proceedings of the ACM Web Conference 2024}, 3833--3843.

\bibitem[{Tang, Rybin, and Chang(2023)}]{tang2023zeroth}
Tang, Z.; Rybin, D.; and Chang, T.-H. 2023.
\newblock Zeroth-order optimization meets human feedback: Provable learning via ranking oracles.
\newblock \emph{arXiv preprint arXiv:2303.03751}.

\bibitem[{Team et~al.(2023)Team, Anil, Borgeaud, Alayrac, Yu, Soricut, Schalkwyk, Dai, Hauth, Millican et~al.}]{team2023gemini}
Team, G.; Anil, R.; Borgeaud, S.; Alayrac, J.-B.; Yu, J.; Soricut, R.; Schalkwyk, J.; Dai, A.~M.; Hauth, A.; Millican, K.; et~al. 2023.
\newblock Gemini: a family of highly capable multimodal models.
\newblock \emph{arXiv preprint arXiv:2312.11805}.

\bibitem[{Von~R{\"u}tte et~al.(2023)Von~R{\"u}tte, Fedele, Thomm, and Wolf}]{von2023fabric}
Von~R{\"u}tte, D.; Fedele, E.; Thomm, J.; and Wolf, L. 2023.
\newblock Fabric: Personalizing diffusion models with iterative feedback.
\newblock \emph{arXiv preprint arXiv:2307.10159}.

\bibitem[{Wallace et~al.(2024)Wallace, Dang, Rafailov, Zhou, Lou, Purushwalkam, Ermon, Xiong, Joty, and Naik}]{wallace2024diffusion}
Wallace, B.; Dang, M.; Rafailov, R.; Zhou, L.; Lou, A.; Purushwalkam, S.; Ermon, S.; Xiong, C.; Joty, S.; and Naik, N. 2024.
\newblock Diffusion model alignment using direct preference optimization.
\newblock In \emph{Proceedings of the IEEE/CVF Conference on Computer Vision and Pattern Recognition}, 8228--8238.

\bibitem[{Wang et~al.(2024)Wang, Spinelli, Wang, Bai, Qin, and Chen}]{wang2024instantstyle}
Wang, H.; Spinelli, M.; Wang, Q.; Bai, X.; Qin, Z.; and Chen, A. 2024.
\newblock Instantstyle: Free lunch towards style-preserving in text-to-image generation.
\newblock \emph{arXiv preprint arXiv:2404.02733}.

\bibitem[{Wu et~al.(2023)Wu, Sun, Zhu, Zhao, and Li}]{wu2023human}
Wu, X.; Sun, K.; Zhu, F.; Zhao, R.; and Li, H. 2023.
\newblock Human Preference Score: Better Aligning Text-to-Image Models with Human Preference.
\newblock arXiv:2303.14420.

\bibitem[{Wu et~al.(2024)Wu, Zhou, Ma, Su, Ma, and Wang}]{wu2024ifadapter}
Wu, Y.; Zhou, X.; Ma, B.; Su, X.; Ma, K.; and Wang, X. 2024.
\newblock Ifadapter: Instance feature control for grounded text-to-image generation.
\newblock \emph{arXiv preprint arXiv:2409.08240}.

\bibitem[{Xu et~al.(2023)Xu, Liu, Wu, Tong, Li, Ding, Tang, and Dong}]{xu2023imagereward}
Xu, J.; Liu, X.; Wu, Y.; Tong, Y.; Li, Q.; Ding, M.; Tang, J.; and Dong, Y. 2023.
\newblock ImageReward: Learning and Evaluating Human Preferences for Text-to-Image Generation.
\newblock arXiv:2304.05977.

\bibitem[{Yang, Feng, and Huang(2024)}]{yang2024emogen}
Yang, J.; Feng, J.; and Huang, H. 2024.
\newblock EmoGen: Emotional Image Content Generation with Text-to-Image Diffusion Models.
\newblock In \emph{Proceedings of the IEEE/CVF Conference on Computer Vision and Pattern Recognition}, 6358--6368.

\bibitem[{Yang et~al.(2023)Yang, Huang, Ding, Lischinski, Cohen-Or, and Huang}]{yang2023emoset}
Yang, J.; Huang, Q.; Ding, T.; Lischinski, D.; Cohen-Or, D.; and Huang, H. 2023.
\newblock Emoset: A large-scale visual emotion dataset with rich attributes.
\newblock In \emph{Proceedings of the IEEE/CVF International Conference on Computer Vision}, 20383--20394.

\bibitem[{Yang et~al.(2024)Yang, Yu, Meng, Xu, Ermon, and Bin}]{yang2024mastering}
Yang, L.; Yu, Z.; Meng, C.; Xu, M.; Ermon, S.; and Bin, C. 2024.
\newblock Mastering text-to-image diffusion: Recaptioning, planning, and generating with multimodal llms.
\newblock In \emph{Forty-first International Conference on Machine Learning}.

\bibitem[{Ye et~al.(2023)Ye, Zhang, Liu, Han, and Yang}]{ye2023ip}
Ye, H.; Zhang, J.; Liu, S.; Han, X.; and Yang, W. 2023.
\newblock Ip-adapter: Text compatible image prompt adapter for text-to-image diffusion models.
\newblock \emph{arXiv preprint arXiv:2308.06721}.

\bibitem[{Yu et~al.(2024)Yu, Lu, Gao, Tan, Yang, Wang, Wu, and Vinitsky}]{yu2024few}
Yu, C.; Lu, H.; Gao, J.; Tan, Q.; Yang, X.; Wang, Y.; Wu, Y.; and Vinitsky, E. 2024.
\newblock Few-shot in-context preference learning using large language models.
\newblock \emph{arXiv preprint arXiv:2410.17233}.

\bibitem[{Zhang, Rao, and Agrawala(2023)}]{zhang2023adding}
Zhang, L.; Rao, A.; and Agrawala, M. 2023.
\newblock Adding conditional control to text-to-image diffusion models.
\newblock In \emph{Proceedings of the IEEE/CVF International Conference on Computer Vision}, 3836--3847.

\bibitem[{Zhang et~al.(2024)Zhang, Wang, Wu, Li, Gao, Zhang, and Wang}]{zhang2024learning}
Zhang, S.; Wang, B.; Wu, J.; Li, Y.; Gao, T.; Zhang, D.; and Wang, Z. 2024.
\newblock Learning Multi-dimensional Human Preference for Text-to-Image Generation.
\newblock In \emph{Proceedings of the IEEE/CVF Conference on Computer Vision and Pattern Recognition}, 8018--8027.

\bibitem[{Zhang et~al.(2023)Zhang, Zhang, Li, Zhao, Karypis, and Smola}]{zhang2023multimodal}
Zhang, Z.; Zhang, A.; Li, M.; Zhao, H.; Karypis, G.; and Smola, A. 2023.
\newblock Multimodal chain-of-thought reasoning in language models.
\newblock \emph{arXiv preprint arXiv:2302.00923}.

\end{thebibliography}

\end{document}